# Improved Monte Carlo tree search formulation with multiple root nodes for discrete sizing optimization of truss structures

Fu-Yao Ko, Katsuyuki Suzuki and Kazuo Yonekura

**Abstract**
This paper proposes a novel reinforcement learning (RL) algorithm using improved Monte Carlo tree search (IMCTS) formulation for discrete optimum design of truss structures. IMCTS with multiple root nodes includes update process, the best reward, accelerating technique, and terminal condition. Update process means that once a final solution is found, it is used as the initial solution for next search tree. The best reward is used in the backpropagation step. Accelerating technique is introduced by decreasing the width of search tree and reducing maximum number of iterations. The agent is trained to minimize the total structural weight under various constraints until the terminal condition is satisfied. Then, optimal solution is the minimum value of all solutions found by search trees. These numerical examples show that the agent can find optimal solution with low computational cost, stably produces an optimal design, and is suitable for multi-objective structural optimization and large-scale structures.

## 1. Introduction

The primary goal of structural optimization is to find minimum weight of the structures satisfying the performance and construction criteria by the specifications and design codes (Venkayya 1978). One of the main issues in the field of structural optimization is sizing optimization of truss structures, where cross-sectional areas of members are design variables. In many practical engineering problems, the variables have to be chosen from the permissible list of discrete values due to the availability of components in standard sizes (Stolpe 2016). The large number of design variables, the highly irregular search space, and the control of a great number of design constraints are major preventive factors in performing optimum design in a reasonable time (Kaveh and Talatahari 2009). The optimization algorithms can be classified into two main categories: traditional mathematical optimization algorithms (Belegundu and Arora 1985) and metaheuristic algorithms (Renkavieski and Parpinelli 2021).

In the past three decades, metaheuristic algorithms have been developed to solve the discrete sizing optimization of truss structures. Some of the most popular methods are:

genetic algorithm (GA) (Rajeev and Krishnamoorthy 1992), ant colony optimization (ACO) (Bland 2001), harmony search (HS) (Lee et al. 2005), and other methods (Hasançebi and Azad 2014; Gholizadeh and Milany 2018). These algorithms provide promising solutions with low computational cost for benchmark and large-scale truss structures. However, the solutions do not satisfy any theoretically defined optimality criteria. Moreover, the quality of the solution heavily relies on the initial solution and parameters of the algorithm. The validation of the parameters should be confirmed only by trial and error (Hayashi and Ohsaki 2021; Hayashi and Ohsaki 2022).

Monte Carlo tree search (MCTS) is a heuristic search algorithm for sequential decision problems. MCTS incrementally and asymmetrically builds a search tree according to the simulation results through random sampling (Browne et al. 2012). MCTS can operate effectively with very little domain-specific knowledge and select optimal decisions in a large search space. However, a huge amount of memory consumption and computational effort is required to construct a search tree. MCTS method has shown exceptional performance in game playing, security, planning and scheduling, chemical synthesis, vehicle routing.

Reinforcement learning (RL) is an interdisciplinary area of machine learning that focuses on the relationship between an agent and its environment. An RL aims to train an action taker called agent to take actions to maximize the cumulative rewards. The RL task is modeled as a Markov decision process (MDP), where the state transition and reward are solely dependent on the state and action taken at the current step. Dynamic programming (DP) is used to find optimal action sequences given a perfect model of the environment as an MDP. The common method is based on the Bellman optimality equation. DP has limited ability in RL since it requires high computational complexity when solving large-scale problems (Sutton and Barto 2018).

The key components of an MDP include agent, environment, state, action, state transition, and reward. For discrete optimum design of truss structures, the state represents the description of the current truss structure. The only action for this problem is to determine the cross-sectional areas of truss members. The environment transitions to a new state after taking an action based on the current state. Reward functions describe how the agent ought to behave. Only the reward for terminal state is considered in this problem.

RL method has achieved remarkable success in game playing, robotics, and scheduling problem. Recently, Q-learning-based RL method has been extensively applied in structural optimization problems (Hayashi and Ohsaki 2020; Zhu et al. 2021; Kupwiwat, Hayashi, and Ohsaki 2024). It is applicable to structural optimization problems because the optimal solution cannot be found beforehand, and RL approach does not require input-output pairs as training data. However, it leads to a high computational cost for training the agent (Hayashi and Ohsaki 2022).

A novel MCTS-based RL algorithm called AlphaTruss (Luo et al. 2022a; Luo et al. 2022b) is proposed to generate optimal truss design considering continuous member size, shape, and topology. It is found that AlphaTruss has a strong generality to be applied to structural optimization problems. However, search tree with exactly one root node is used, which is expected to trap into the local optimal solution. Moreover, this method is not verified by benchmark problems with discrete variables and not applied to multi-objective structure optimization and large-scale problems.

In this paper, an RL algorithm using improved Monte Carlo tree search (IMCTS) formulation with multiple root nodes is developed for discrete sizing optimization of truss structures. This algorithm incorporates update process, the best reward, accelerating technique, and terminal condition. The agent is trained to minimize the weight of the truss under multiple constraints. The proposed algorithm is tested on several benchmark truss problems by comparing the results from the literature previously been analyzed by metaheuristic algorithms. The results show that IMCTS formulation can find optimal solutions without parameter tuning. Also, this algorithm can solve multi-objective structural optimization and practical engineering problems.

## 2. Preliminary knowledge: mathematical model, MCTS, and MDP for discrete sizing optimization of truss structures

In this section, a mathematical model is first detailed for sizing optimization of truss structures with discrete variables. Then, MCTS and MDP applied to discrete truss optimization are depicted because they are directly related to the proposed algorithm in this paper. Finally, the difference between the DP, the AlphaTruss, and the IMCTS formulation for tree structures is presented.

2.1 Mathematical model for discrete optimum design of truss structures

A structural optimization problem with discrete design variables can be formulated as a nonlinear programming problem with multiple nonlinear constraints associated with structural behavior. In discrete sizing optimization problems, the main task is to select the cross-sectional areas of the truss members. All of them are selected from a permissible list of standard sections. The optimization problem aims to minimize the weight of the truss while stress and displacement constraints are fulfilled. The optimization design problem for discrete variables can be expressed as follows:

$$\text{Find} \quad \begin{aligned} &\boldsymbol{X} = (X_1, X_2, \dots, X_i, \dots, X_g), \\ &X_i \in \mathbf{D}, \mathbf{D} = \{d_1, d_2, \dots, d_h, \dots, d_b\} \end{aligned} \quad (1a)$$

$$\text{Minimize} \quad W(X) = \rho \sum_{i=1}^{g} \left( X_i \sum_{j=1}^{m_i} L_{i,j} \right) \tag{1b}$$

where $X$ is the vector containing the design variables, which are selected from a list of available discrete values; $X_i$ is the cross-sectional area of the members belonging to group $i$; $g$ is the number of member groups (design variables); $\mathbf{D}$ is the list including all available discrete values arranged in ascending sequences; $d_h$ is the element of a list $\mathbf{D}$; $h$ is the index of permissive discrete variables; $b$ is the number of available sections; $W(X)$ is the weight of the truss; $\rho$ is the material density; $m_i$ is the number of members in the group $i$; and $L_{i,j}$ is the length of the member $j$ in the group $i$. The constraints can be formulated as follows:

$$\text{Subject to} \quad \begin{aligned} \mathbf{K}(X)\mathbf{u} &= \mathbf{F} & (2a) \\ \varepsilon_{i,j} &= \mathbf{B}_{i,j}\mathbf{u}_{i,j} & (2b) \\ \sigma_{i,j} &= \frac{E}{L_{i,j}} \varepsilon_{i,j} & (2c) \\ \sigma_{\min} &\leq \sigma_{i,j} \leq \sigma_{\max}, i=1,2,\ldots,g, j=1,2,\ldots,m_i & (2d) \\ \delta_{\min} &\leq \delta_k \leq \delta_{\max}, k=1,2,\ldots,c & (2e) \end{aligned}$$

where $\mathbf{K}(X)$ is the global stiffness matrix of the structure; $\mathbf{u}$ is a vector containing the displacements of the non-suppressed nodes of the truss; $\mathbf{F} = (F_1^x, F_1^y, F_1^z, F_2^x, F_2^y, F_2^z, \ldots, F_k^x, F_k^y, F_k^z, \ldots)$ is a vector containing the applied forces at these nodes; $F_k^x$, $F_k^y$, and $F_k^z$ are the $x$, $y$, and $z$ components of applied force acting on node $k$; $\varepsilon_{i,j}$ is the elongation of the member; $\mathbf{B}_{i,j}$ is defined as $\mathbf{B}_{i,j} = (-\mathbf{e}_{i,j}^{tr} \quad \mathbf{e}_{i,j}^{tr})$; $\mathbf{e}_{i,j}$ is a unit vector along the member so that it points from local node 1 to local node 2; $\mathbf{u}_{i,j} = (\mathbf{u}_{i,j}^1 \quad \mathbf{u}_{i,j}^2)^{tr}$ is a vector including the displacements of the end points of the member; tr in the superscript of the given vector is the transpose operator; $E$ is the modulus of elasticity; and $c$ is the number of structural joints. From Equations (2d) and (2e), the stress $\sigma_{i,j}$ in each member of the truss is compared with $\sigma_{\min}$ and $\sigma_{\max}$. The displacement $\delta_k$ of node $k$ of the truss is compared with $\delta_{\min}$ and $\delta_{\max}$. $\sigma_{\min}$ and $\sigma_{\max}$ refer to the lower and upper bound of the stress. $\delta_{\min}$ and $\delta_{\max}$ refer to the lower and upper bound of the displacement.

Fig. 1 shows an example of 2-bar planar truss to introduce the notation mentioned above. This example includes 2 member groups and 3 structural joints. $x_k$, $y_k$, and $z_k$ are the coordinates of node $k$ in the $x$, $y$, and $z$ directions, respectively. $R_k^x$, $R_k^y$, and $R_k^z$ are the $x$, $y$, and $z$ components of reaction force applied to node $k$.

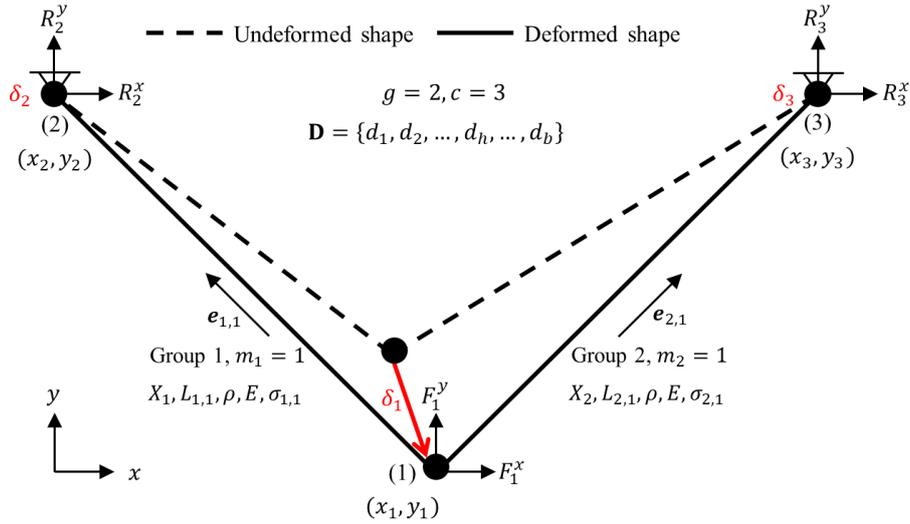

Fig. 1 An example of 2-bar planar truss for truss optimization problem

2.2 MCTS methodology for structural optimization

MCTS is an iterative, guided, random best-first tree search algorithm effectively applied to truss sizing optimization problems with discrete variables. It uses a search tree to model the problem. MCTS starts with a search tree containing only the root node, expands the search tree based on random sampling, and selects the most promising moves. The search tree for structural optimization problem is shown in Fig. 2. Four steps shown in Fig. 3 are applied in each iteration until the predefined computational budget is reached. The four steps are elaborated in the following:

- Selection: Starting at the root node, the most promising child node is selected recursively to descend through the tree following a tree policy until a leaf node is reached. A leaf node represents unvisited child node or terminal node.
- Expansion: All child nodes are added to expand the search tree when the selected leaf node is not a terminal node.
- Simulation: After adding the new nodes to the tree, one node is randomly chosen for simulation. Then, a simulation is run according to the default policy until reaching a terminal node to produce a result.
- Backpropagation: The simulation result is backpropagated from the selected leaf node to the root node. Statistics are updated for selected node during the selection, and visit counts are increased.

In the selection step, the upper confidence bound (UCB) used to select the best child node (Kocsis and Szepesvári 2006) is defined as follows:

$$U_I = \frac{\sum G_{n_I}}{n_I} + C\sqrt{\frac{\ln N}{n_I}} \qquad (3)$$

where $I$ is the index of the node; $U_I$, $\sum G_{n_I}$, and $n_I$ are the UCB, the sum of the simulation results, and the number of simulations conducted for node $I$, respectively; $N$ is the total number of simulations executed from the parent node; and $C$ is a constant parameter to control the trade-off between exploration and exploitation, empirically set to $\sqrt{2}$ in this study. The term $\sum G_{n_I}/n_I$ is called the average reward in Equation (3). In the backpropagation step, $n_I$ increases by 1 for all nodes along the path, which is formulated as follows:

$$n_I \leftarrow n_I + 1 \tag{4}$$

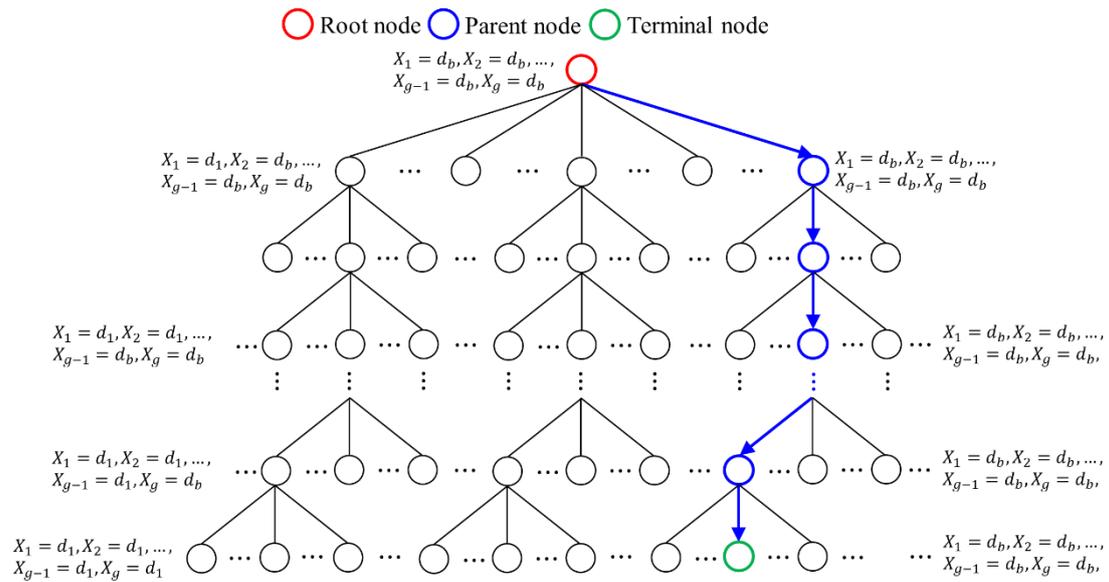

Fig. 2 Search tree for structural optimization is constructed for MCTS

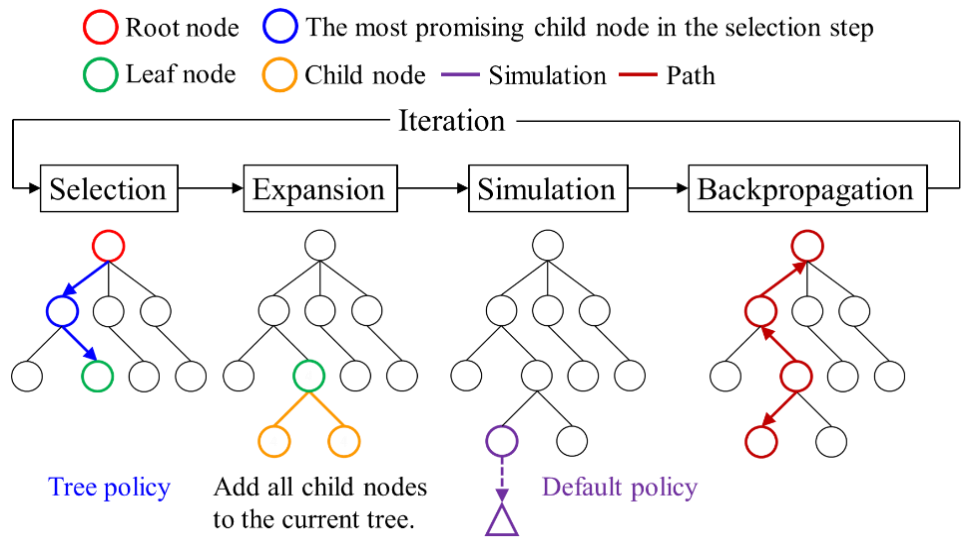

Fig. 3 Four steps of the MCTS algorithm (Browne et al. 2012)

2.3. MDP for structural optimization

Fig. 4 illustrates the MDP used in this research. At each time step $t$, the agent observes the state $s_t$ as numerical data of nodes and members representing the truss structure, as shown in Fig. 4(a). Then, it takes the action $a_t$ to determine the cross-sectional areas of truss members. Fig. 4(b) shows that the cross-sectional area $X_1$ is changed from $d_b$ to $d_h$. When the agent transitions to a new state, it receives a reward $r_{t+1}$ computed from the truss optimization problem, as shown in Fig. 4(c).

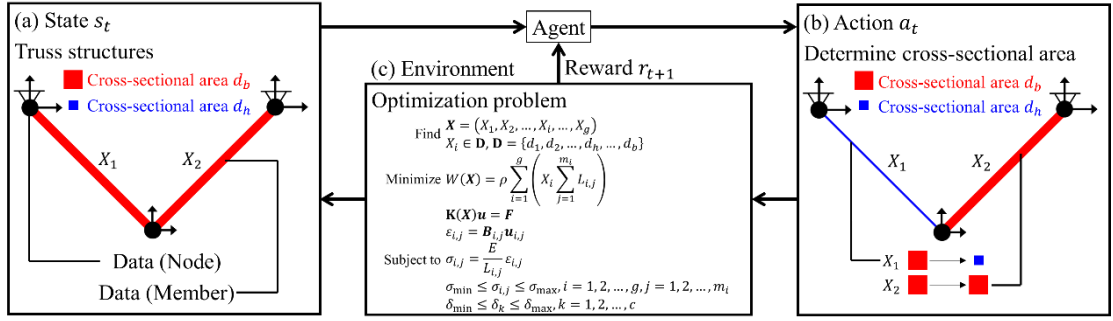

Fig. 4 Proposed MDP for sizing optimization of truss structures

2.3.1 State

Numerical data of a truss structure is suitable for describing the state. Therefore, a state is expressed as a set of numerical data of nodes $\hat{\mathbf{v}} = \{\mathbf{v}_1, \mathbf{v}_2, \ldots, \mathbf{v}_k, \ldots, \mathbf{v}_c\}$ and that of members $\hat{\mathbf{w}} = \{\mathbf{w}_{1,1}, \mathbf{w}_{1,2}, \ldots, \mathbf{w}_{i,j}, \ldots, \mathbf{w}_{g,m_g}\}$.

(1) Node feature vector

The necessary node features include the nodal coordinates, the applied forces, and the reaction forces. Therefore, the feature vector of node $k$ is constructed as follows:

$$\mathbf{v}_k = \left(x_k, y_k, z_k, F_k^x, F_k^y, F_k^z, R_k^x, R_k^y, R_k^z\right) \tag{5}$$

(2) Member feature vector

The necessary member features include the node numbers of the end points of the member, length, the modulus of elasticity, index to confirm when cross-sectional area is determined, and cross-sectional area. Hence, the feature vector of the member $j$ in the group $i$ is represented as follows:

$$\mathbf{w}_{i,j} = \left(H_{i,j}^1, H_{i,j}^2, L_{i,j}, E, M_{X,i}, X_i\right) \tag{6}$$

where $H_{i,j}^1$ and $H_{i,j}^2$ are the node numbers of both ends of the member; and $M_{X,i}$ is used to check when cross-sectional area of the member group $i$ is determined. $M_{X,i} = 1$ when $X_i$ is not determined, and $M_{X,i} = 0$ otherwise.

States in the MDP model shown in Fig. 5 can be classified into three types: initial state (red circle), intermediate state (blue circle), and terminal state (green circle). Initial state means that all design variables are not determined. Only one initial state is included in the MDP model. Intermediate state means that some design variables are determined, and some are not. Terminal state means that all design variables are determined. $s_0$, $s_{l,q}$ ($l = 1, 2, ..., g-1$), and $s_{g,q}$ denote the initial state, the intermediate state, and the terminal state, respectively. $l$ is the layer number. $q$ is the index of state for layer $l$. The mathematical expressions of $s_0$, $s_{l,q}$, and $s_{g,q}$ are written as follows:

$$\text{State } s_0: M_{X,1} = M_{X,2} = \cdots = M_{X,g} = 1 \tag{7a}$$

$$\begin{array}{l}\text{State } s_{l,q}: M_{X,1} = M_{X,2} = \cdots = M_{X,i} = 0 \\ \text{and } M_{X,i+1} = M_{X,i+2} = \cdots = M_{X,g} = 1 \ (i = l)\end{array} \tag{7b}$$

$$\text{State } s_{g,q}: M_{X,1} = M_{X,2} = \cdots = M_{X,g} = 0 \tag{7c}$$

2.3.2 Action

For discrete truss optimization, action is defined as determination of a cross-sectional area from a permissible list. Actions in the MDP model shown in Fig. 5 can be divided into two categories: $a_{l,f}$ (black circle) and $a_T$ (orange circle). Action $a_{l,f}$ ($l = 0, 1, 2, ..., g-1$) for initial and intermediate state is described as follows:

$$\text{Action } a_{l,f}: M_{X,i}: 1 \to 0 \text{ and } A_i: d_b \to d_h \ (i = l+1, f = h) \tag{8}$$

where $f$ is the index of action for layer $l$. Action $a_T$ for terminal state $T$ means that the state is not changed when it takes the action $a_T$. The action spaces of $s_0$, $s_{l,q}$, and $s_{g,q}$ are as follows:

$$\mathcal{A}(s_0) = \{a_{0,1}, a_{0,2}, ..., a_{0,f}, ..., a_{0,b}\} \tag{9a}$$

$$\mathcal{A}(s_{l,q}) = \{a_{l,1}, a_{l,2}, ..., a_{l,f}, ..., a_{l,b}\} \tag{9b}$$

$$\mathcal{A}(s_{g,q}) = \{a_T\} \tag{9c}$$

Based on Equations (9a) and (9b), the number of elements in $\mathcal{A}(s_0)$ and $\mathcal{A}(s_{l,q})$ is the number of available sections $b$.

2.3.3 Reward

Reward is evaluated only for terminal state. In other words, reward for initial and intermediate state is 0. When the truss violates stress or displacement constraints, reward of 0 is assigned. Otherwise, the reward $r_T$ is calculated as follows:

$$r_T = (\alpha/W_T)^2 \tag{10}$$

where $r_T$ is the reward for terminal state; $\alpha$ is defined as the weight of the truss with cross-sectional areas of all groups equal to the smallest element in a list **D**, i.e., $d_1$, which is called the minimum weight in this study; and $W_T$ is the weight of the truss for terminal state. $r_T$ is a dimensionless quantity used to minimize the weight of the truss under various structural constraints. The MDP model shown in Fig. 5 can be represented by a tree structure with $(g + 1)$ layers, including state (Section 2.3.1), action (Section 2.3.2), and state transition. In Fig. 5, state, action, and state transition are represented by hollow circle, solid circle, and purple arrow, respectively.

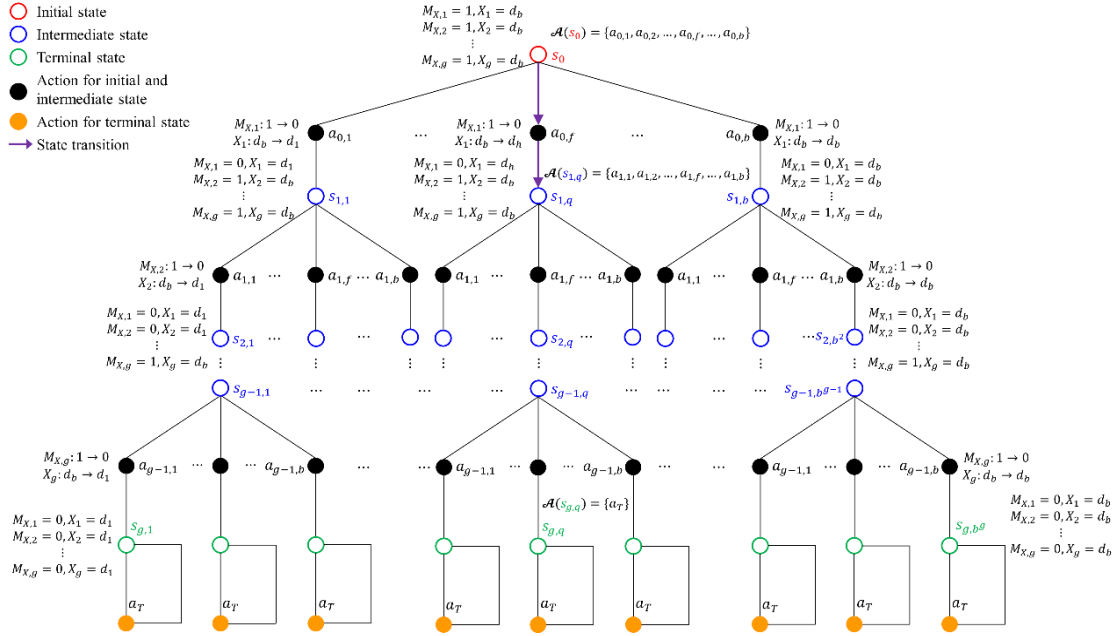

Fig. 5 MDP model represented by a tree structure

2.4. Difference between DP, AlphaTruss, and IMCTS formulation

Fig. 6(a), (b), and (c) shows the tree structure of DP, AlphaTruss, and IMCTS formulation, respectively. In Fig. 6(b) and (c), the state and action are denoted by hollow circle and solid line, respectively. As shown in Fig. 6(a), DP uses a complete model of the environment to find an optimal policy. This method goes through all possible states and actions, which is called full backup. In order to deal with such situation, AlphaTruss shown in Figs. 6(b) and 7 uses the search tree with single root node to find the optimal solution. Action and action space for initial and intermediate state are described in Section 2.3.2. When final state $\bar{s}$ is found by search tree, design variable vector $\bar{X} = (\overline{X_1}, \overline{X_2}, ..., \overline{X_i}, ..., \overline{X_g})$ is determined. However, it has a high possibility to trap into the local optimal solution. In order to solve this issue, IMCTS formulation with multiple root nodes is developed in this study. The working of the IMCTS formulation is explained in detail in the next section.

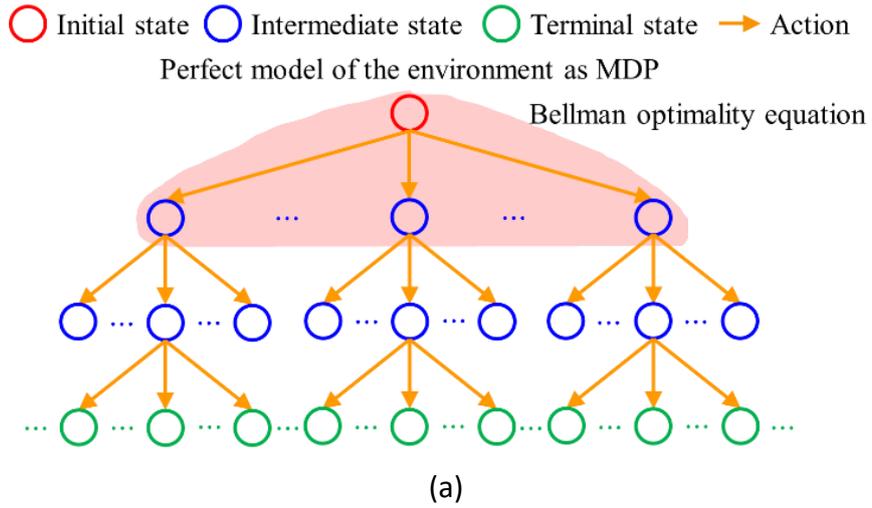

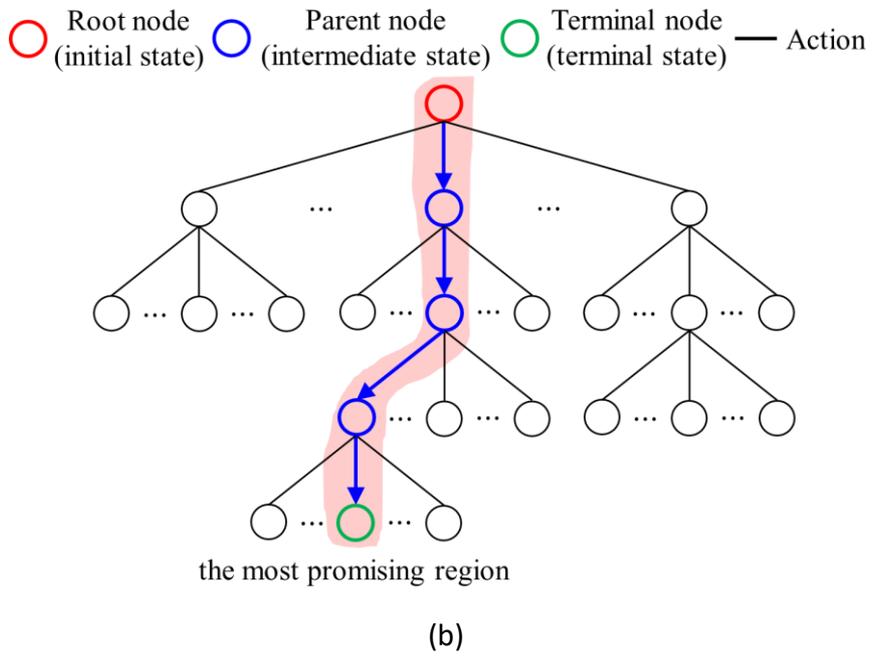

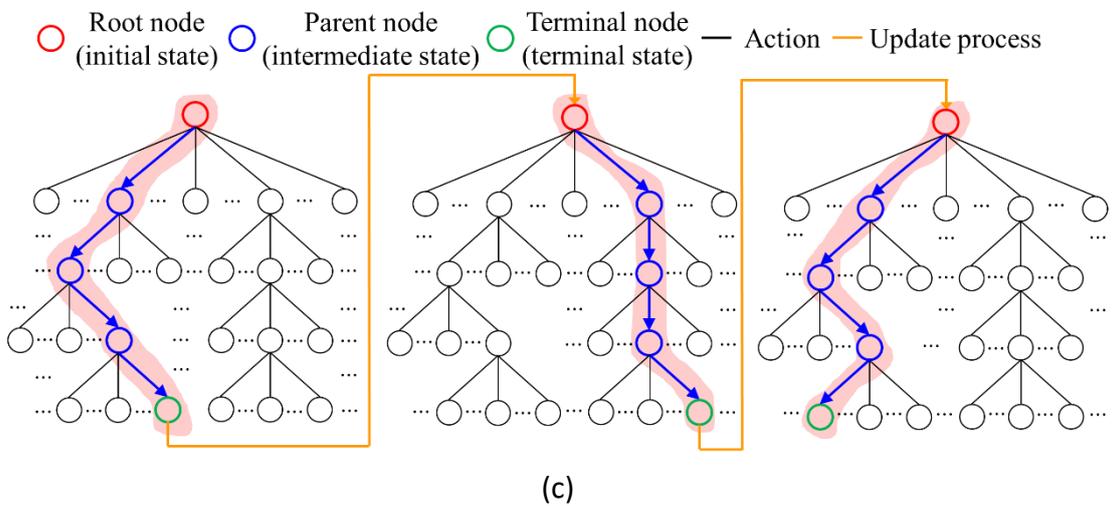

**Fig. 6** Tree structure for (a) DP, (b) AlphaTruss, and (c) IMCTS formulation

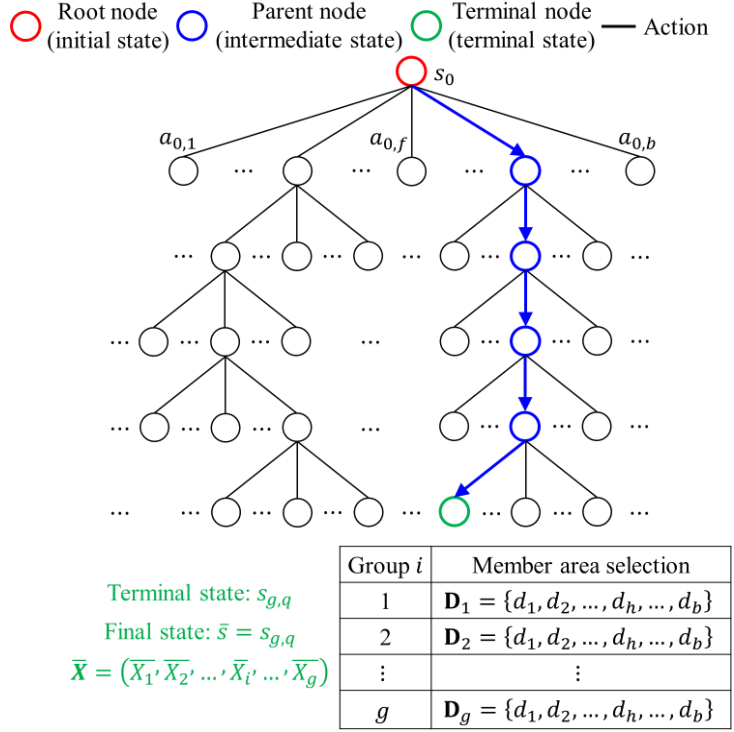

**Fig. 7** Schematic diagram of the AlphaTruss using search tree with single root node

## 3. An RL algorithm using IMCTS formulation with multiple root nodes for discrete optimum design of truss structures

The MCTS and MDP introduced in the previous section can be integrated for discrete structural optimization. This section describes the RL algorithm using IMCTS formulation with multiple root nodes. IMCTS formulation incorporates update process, the best reward, accelerating technique, and terminal condition. The concepts are briefly explained as follows.

3.1 MCTS methodology for sizing optimization of truss structures

In this study, a search tree is constructed in each round for structural optimization problems. Each node represents the state of the current truss structure. $s_0^p$, $s_{l,q}^p$, and $s_{g,q}^p$ denote the initial state, the intermediate state, and the final state in round $p$, respectively. The mathematical expressions of $s_0^p$, $s_{l,q}^p$, and $s_{g,q}^p$ are written as follows:

$$\text{State } s_0^p: M_{X,1}^p = M_{X,2}^p = \cdots = M_{X,g}^p = 1 \tag{11a}$$

$$\text{State } s_{l,q}^p: M_{X,1}^p = M_{X,2}^p = \cdots = M_{X,i}^p = 0$$
$$\text{and } M_{X,i+1}^p = M_{X,i+2}^p = \cdots = M_{X,g}^p = 1 \ (i = l) \tag{11b}$$

$$\text{State } s_{g,q}^p: M_{X,1}^p = M_{X,2}^p = \cdots = M_{X,g}^p = 0 \tag{11c}$$

where $M_{X,i}^p$ is used to confirm when cross-sectional area of the member group $i$ in round $p$ is determined. The root node corresponds to the initial state $s_0^p$ in round $p$. Nodes with terminal states $s_{g,q}^p$ are called terminal nodes. Each edge represents the action, which will be further explained in Section 3.3. MCTS starts with a single root node and iteratively builds a partial search tree, as shown in Fig. 8. This process is repeated until reaching the predefined computational budget. The four steps are described in detail in Section 2.2. It is noting that state values and visit counts of all added child nodes are initialized in the expansion step.

In the selection step, the UCB for IMCTS formulation based on Eq. 3 in Section 2.2 is defined as follows:

$$U_I = V_I + C \sqrt{\frac{\ln N}{n_I}} \tag{12}$$

where $V_I$ is the estimate of state value for node $I$. In Eq. 12, state value instead of average reward is used to calculate the UCB. Because the purpose of this method is to guarantee an optimal solution path, the best reward used in the backpropagation step is defined as follows:

$$V_I \leftarrow \max(V_I, G_{\tau_N}) \tag{13}$$

where $G_{\tau_N}$ is the simulation result of the path $\tau_N$ from the considered root node or parent node. The best reward is the maximum value between the current state value and the simulation result. Equation (4) is also used in the backpropagation step.

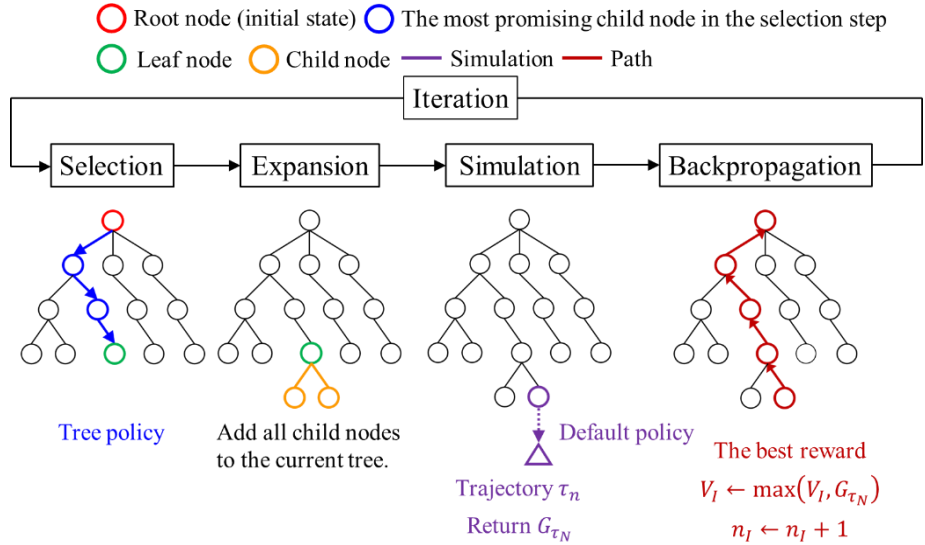

**Fig. 8** The four steps for IMCTS formulation considering the best reward

3.2. Policy improvement

The policy improvement process is shown in Fig. 9. MCTS starts from the root node

(red circle) and continues executing these four steps. After maximum number of iterations for root node is reached, a child node with the largest estimate of the state value (blue circle) is selected. This node is regarded as the parent node for the next policy improvement. Then, MCTS starts at the parent node (green circle), conducts the four steps iteratively until reaching maximum number of iterations, and the child node with the largest estimate of state value (blue circle) is chosen. After many policy improvement steps, terminal node (orange circle) is reached. The state of the terminal node is called the final state $\overline{s^p}$ in round $p$. The final weight $\overline{W^p}$ in round $p$ is defined as the weight of the truss for final state.

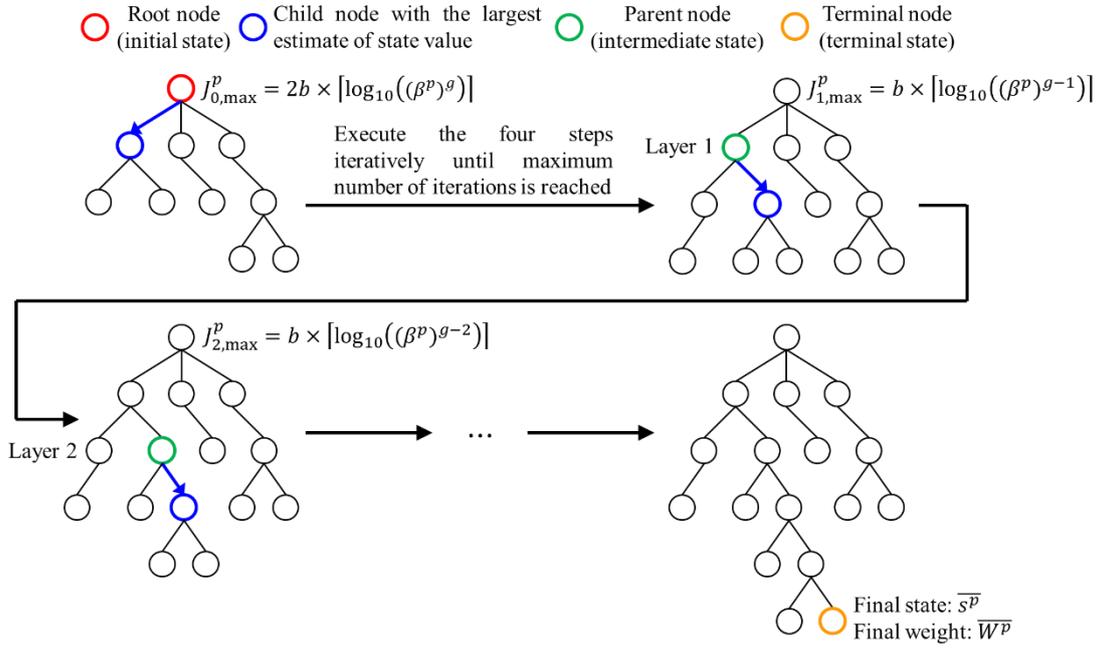

**Fig. 9** Policy improvement process for IMCTS formulation

### 3.3. Update process

Once a final state $\overline{s^p}$ (green circle) is found by search tree, design variable vector $\overline{X^p} = \left(\overline{X_1^p}, \overline{X_2^p}, \dots, \overline{X_i^p}, \dots, \overline{X_g^p}\right)$ in round $p$ is determined. Then, $\overline{X^p}$ is used as the design variable vector $X_0^{p+1} = \left(\left(X_1^{p+1}\right)_0, \left(X_2^{p+1}\right)_0, \dots, \left(X_i^{p+1}\right)_0, \dots, \left(X_g^{p+1}\right)_0\right)$ for initial state $s_0^{p+1}$ (red circle) in round $(p+1)$. It is called an update process (orange line) in this study, as shown in Fig. 10. Since the starting point of a path is the maximum value, all design variables for initial state $s_0^1$ in round 1 is equal to the largest element in a list **D**, i.e., $d_b$. The center of the search space needs to be the design variable vector for initial state. For this purpose, a list $\mathbf{D}_i^p = \left\{\left(d_i^p\right)_1, \dots, \left(d_i^p\right)_h, \dots, \left(d_i^p\right)_\mu, \dots, \left(d_i^p\right)_{\beta_p}\right\}$ with member areas belonging to **D** is defined, where $\left(d_i^p\right)_\mu$ is the median of the list $\mathbf{D}_i^p$ and equal to $\left(X_i^p\right)_0$; and $\beta^p$ is the number of elements in the list $\mathbf{D}_i^p$. Tables in Fig. 10 present the list $\mathbf{D}_i^p$ of the group $i$ in round $p$. When considering update process, action $a_{l,f}^p$ ($l = 0, 1, 2, \ldots, g - 1$) for initial and intermediate state is described as follows:

$$\text{Action } a_{l,f}^p: M_{X,i}^p: 1 \to 0 \text{ and } X_i^p: \left(X_i^p\right)_0 \to \left(d_i^p\right)_h \ (i = l + 1, f = h) \tag{14}$$

where $X_i^p$ is the cross-sectional area of the group $i$ in round $p$. The action spaces of $s_0^p$ and $s_{l,q}^p$ are as follows:

$$\mathcal{A}(s_0^p) = \{a_{0,1}^p, a_{0,2}^p, \ldots, a_{0,f}^p, \ldots, a_{0,\beta^p}^p\} \tag{15a}$$
$$\mathcal{A}(s_{l,q}^p) = \{a_{l,1}^p, a_{l,2}^p, \ldots, a_{l,f}^p, \ldots, a_{l,\beta^p}^p\} \tag{15b}$$

Based on Equations (15a) and (15b), the number of elements in $\mathcal{A}(s_0^p)$ and $\mathcal{A}(s_l^p)$ is equal to $\beta^p$.

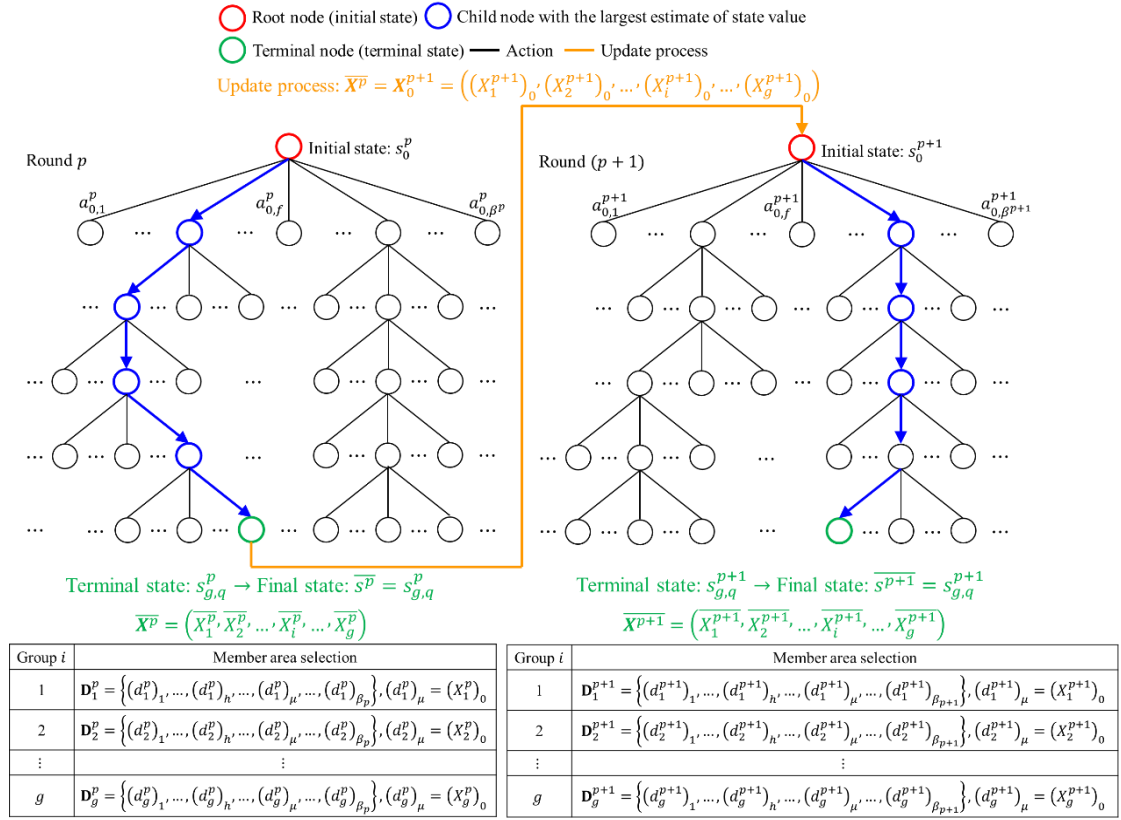

**Fig. 10** Schematic diagram of the IMCTS formulation with update process.

3.4 Accelerating technique

The width of search tree is not changed as the update process proceeds when accelerating technique is not considered, as shown in Fig. 11(a). In other words, $\beta^p$ is a constant for each round, which is formulated as follows:

$$\beta^1 = \beta^2 = \cdots = \beta^p = \cdots = \begin{cases} b \text{ if } b \text{ is odd number} \\ b+1 \text{ if } b \text{ is even number} \end{cases} \quad (16)$$

From Equation (16), it is seen that $\beta^p$ is based on the number of available variables $b$.

Accelerating technique is developed by decreasing the width of search tree during the update process, as shown in Fig. 11(b). Three types of accelerating techniques are considered in this study: geometric decay, linear decrease, and step reduction.

In the geometric decay, $\beta^p$ is computed as follows:

$$\beta^1 = \begin{cases} b \text{ if } b \text{ is odd number} \\ b+1 \text{ if } b \text{ is even number} \end{cases} \quad (17a)$$

$$\varphi^p = \beta^1 \times \gamma_{geo}^{\left\lceil \frac{p-1}{\epsilon_{geo}} \right\rceil} \quad (17b)$$

$$\phi^p = \lfloor \varphi^p \rfloor \quad (17c)$$

$$\omega^p = \begin{cases} \phi^p \text{ if } \phi^p \text{ is odd number} \\ \phi^p + 1 \text{ if } \phi^p \text{ is even number} \end{cases} \quad (17d)$$

$$\beta^p = \max(3, \omega^p) \ (p > 1) \quad (17e)$$

In Equation (17a), $\beta^1$ is based on the number of available sections $b$. In order to reduce the computational time, $\beta^p$ is decreasing geometrically when more round is reached. Therefore, Equation (17b) is used to satisfy this requirement, where $\gamma_{geo}$ and $\epsilon_{geo}$ are constant parameters for geometric decay to adjust $\beta^p$ in each round; and $\left\lceil \frac{p-1}{\epsilon_{geo}} \right\rceil$ is the least integer greater than or equal to $\frac{p-1}{\epsilon_{geo}}$. In this study, $\gamma_{geo}$ and $\epsilon_{geo}$ are set to 0.5 and 3, respectively. Equation (17c) is utilized to ensure that $\beta^p$ is an integer, where $\lfloor \varphi^p \rfloor$ is the greatest integer less than or equal to $\varphi^p$. Equation (17e) is adopted because the minimum value of $\beta^p$ needs to be 3. The number of elements from the left and right sides of $(d_i^p)_\mu$ should be the same because the same possibility of searching both directions is necessary. Hence, Equations (17a) and (17d) are used to ensure that $\beta^p$ is the odd number.

In the linear decrease, $\beta^p$ is calculated as follows:

$$\beta^1 = \begin{cases} \lceil 0.5b \rceil \text{ if } b \text{ is odd number} \\ 0.5b + 1 \text{ if } b \text{ is even number} \end{cases} \quad (18a)$$

$$\zeta^p = \beta^1 - \gamma_{lin}(p-1) \quad (18b)$$

$$\beta^p = \max(3, \zeta^p) \ (p > 1) \quad (18c)$$

In Equation (18b), $\beta^p$ is decreasing linearly when more round is reached, where $\gamma_{lin}$ is a constant parameter for linear decrease to adjust $\beta^p$ in each round. In this study, $\gamma_{lin}$ is set to 2.

In the step reduction, $\beta^p$ is given as follows:

$$\beta^1 = \begin{cases} \lceil 0.5b \rceil \text{ if } b \text{ is odd number} \\ 0.5b + 1 \text{ if } b \text{ is even number} \end{cases} \quad (19a)$$

$$\vartheta^p = \beta^1 - \gamma_{red} \left\lfloor \frac{p-1}{\epsilon_{red}} \right\rfloor \quad (19b)$$

$$\beta^p = \max(3, \vartheta^p) \ (p > 1) \quad (19c)$$

In Equation (19b), $\beta^p$ is a step function of round number $p$, where $\gamma_{red}$ and $\epsilon_{red}$ are constant parameters for step reduction to adjust $\beta^p$ in each round. In this study, $\gamma_{red}$ and $\epsilon_{red}$ are set to 2 and 3, respectively.

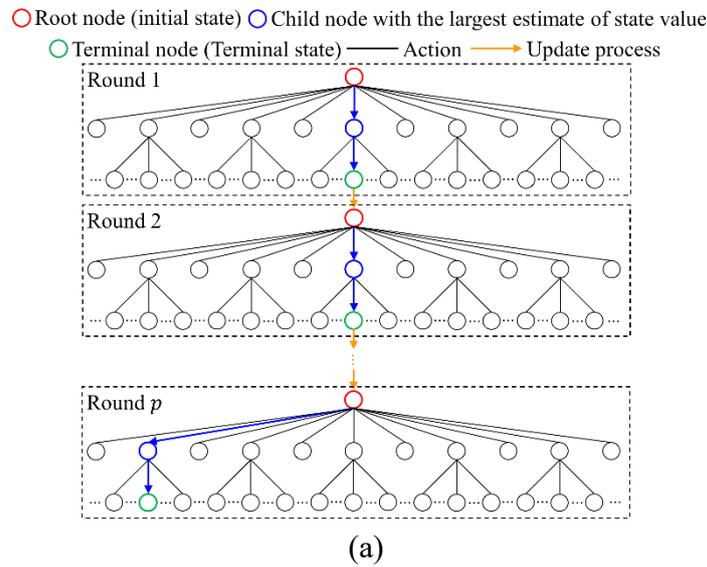

(a)

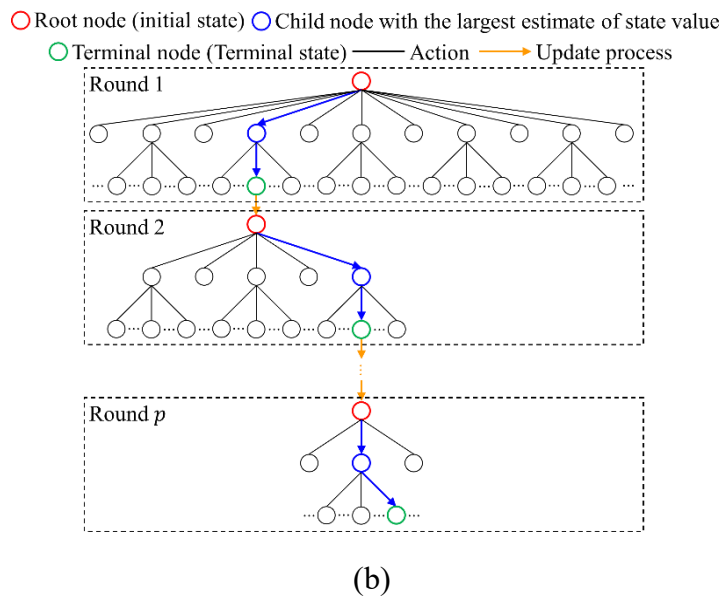

(b)

**Fig. 11** IMCTS formulation (a) without; (b) with accelerating technique.

When performing more and more policy improvement steps, maximum number of iterations needs to be decrease geometrically. Hence, maximum number of iterations for root node $J^p_{0,\max}$ and parent node $J^p_{l,\max}$ for layer $l$ ($l = 1, 2, \ldots, g-1$) is expressed as follows:

For root node (initial state) $\quad\quad\quad J^p_{0,\max} = 2b \times \lceil \log_{10}((\beta^p)^g) \rceil \quad\quad$ (20a)

For parent node (intermediate state) $\quad J^p_{l,\max} = b \times \lceil \log_{10}((\beta^p)^{g-l}) \rceil \quad\quad$ (20b)

### 3.5. Terminal condition

Update process presented in Section 3.3 is carried out multiple rounds until the terminal condition is satisfied. For terminal condition, a list **S** is defined to store the final weight $\overline{W^p}$ in round $p$. $\overline{W^0}$ is defined as the weight of the truss with all design variables set to $d_b$, which is called the maximum weight in this study. Improvement factor $\eta$ and counter $\theta$ are defined to ensure convergence of the algorithm. The improvement factor is defined as follows:

$$\eta = |(\overline{W^p} - \min(\mathbf{S}))/\min(\mathbf{S}) \times 100\%| \quad\quad (21)$$

where min (**S**) is the smallest element in a list **S**. Before execution of an algorithm, $\overline{W^0}$ is in a list **S**, and $\theta$ is set to 0. At the end of the round, the improvement factor $\eta$ is calculated, and then $\overline{W^p}$ is inserted in a list **S**. When $\eta < \eta_{\min}$, $\theta$ increases by 1. When $\theta \geq \theta_{\max}$, the algorithm terminates. $\eta_{\min}$ and $\theta_{\max}$ are the critical value of improvement factor and the maximum number of counters for termination, respectively. In this study, $\eta_{\min}$ and $\theta_{\max}$ are set to 0.01% and 3. At this time, min (**S**) is the optimal solution of the optimization problem in Section 2.1. The flowchart of the terminal condition is shown in Fig. 12. The flowchart of the IMCTS formulation is illustrated in Fig. 13. The pseudo-code for the IMCTS formulation is shown in Fig. 14.

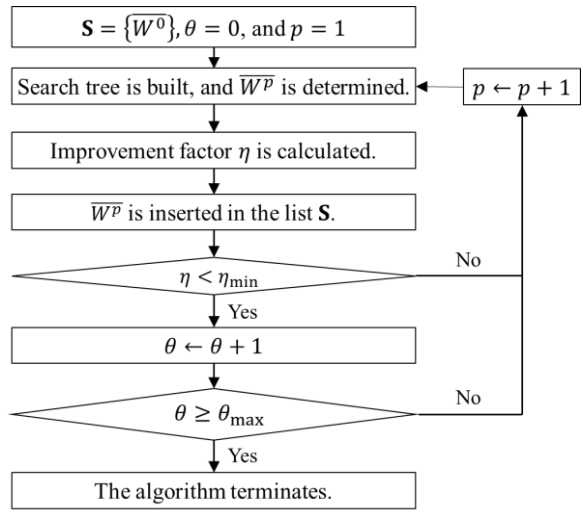

**Fig. 12** Terminal condition for IMCTS formulation.

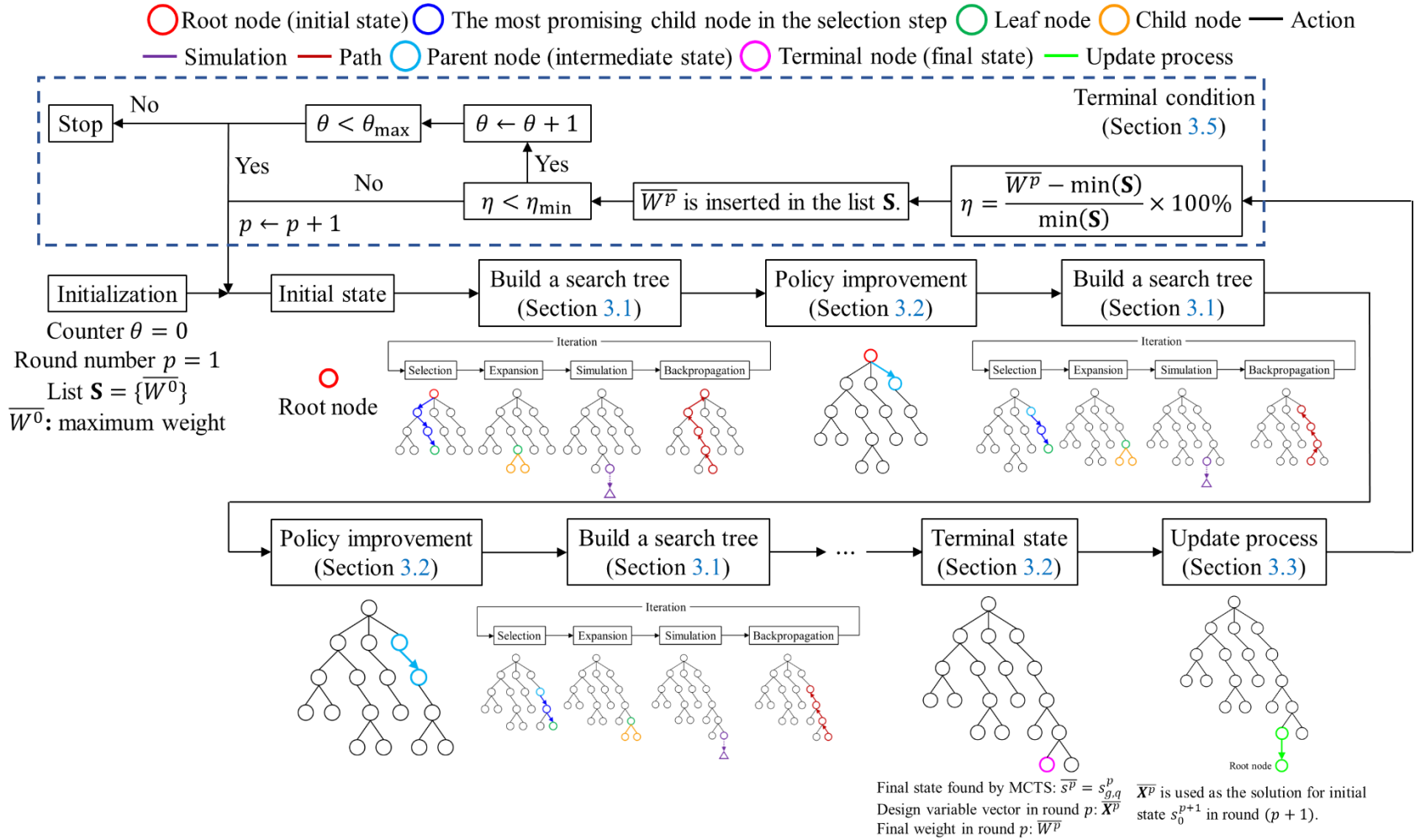

**Fig. 13** Flowchart of the proposed IMCTS formulation.

Algorithm 1: IMCTS formulation
---
**1 Initialize** Round number $p = 1$
**2 Initialize** Counter $\theta = 0$ and maximum number of counters for termination $\theta_{\max} = 3$
**3 Initialize** Critical value of improvement factor $\eta_{\min} = 0.01$
**4 Initialize** List $\mathbf{S} = \{\overline{W^0}\}$ and maximum weight $\overline{W^0}$
**5 Initialize** Design variable vector for initial state $s_0^1$ is $\mathbf{X}_0^1 = (d_b, d_b, \ldots, d_b)$
**6 While** $\theta \leq \theta_{\max}$ **do**
7    Start from a search tree with only root node (initial state $s_0^p$)
8    Define list $\mathbf{D}_i^p$ with member areas belonging to $\mathbf{D}$
9    **While** Terminal node is reached **do**
10       **While** Maximum number of iterations is reached **do**
11          Execute the four steps from the root node or parent node
12          Policy improvement: select a child node with the largest estimate of state value
13          This child node is regarded as the parent node
14       The final state $\overline{s^p}$ is the state of the terminal node
15       Determine design variable vector $\overline{\mathbf{X}^p}$
16       Determine the final weight $\overline{W^p}$
17       Calculate improvement factor $\eta = |(\overline{W^p} - \min(\mathbf{S}))/\min(\mathbf{S}) \times 100\%|$
18       **if** $\eta \leq \eta_{\min}$ **then**
19          $\theta \leftarrow \theta + 1$
20       $\overline{W^p}$ is inserted in list $\mathbf{S}$
21       Update process $\mathbf{X}_0^{p+1} = \overline{\mathbf{X}^p}$
22       $p \leftarrow p + 1$
23 $\min(\mathbf{S})$ is the optimal solution of the optimization problem in Section 2.

**Fig. 14** Pseudo-code for the IMCTS formulation

### 3.6 Characteristics of AlphaTruss and IMCTS formulation

Table 1 summarizes the characteristics of AlphaTruss and IMCTS formulation. It is found that AlphaTruss and IMCTS formulation are applied to continuous and discrete variables, respectively. Moreover, tree structure, reward function, backpropagation step, accelerating technique, and terminal condition are different from these two methods.

**Table 1.** Comparisons of AlphaTruss and IMCTS formulation.

|  | AlphaTruss | IMCTS formulation |
|---|---|---|
| Variable | Continuous variable | Discrete variable |
| Large-scale structures | No | Yes |
| Multi-objective | No | Yes |
| Reward function | $r_T = \lambda/W_T^2$ | $r_T = (\alpha/W_T)^2$ |
| Algorithm | Two-stage | Update process |
| Search tree | Single root node | Multiple root nodes |
| Backpropagation step | The average reward | The best reward |
| Accelerating technique | None | 1. Decrease the width of search tree (1)Geometric decay (2)Linear decrease (3)Step reduction 2. Reduce maximum number of iterations |
| Terminal condition | 1. Generate a structure with a higher weight than the previous round 2. Reach 25 rounds | Improvement factor $\eta$ and counter $\theta$ are defined. When $\eta < \eta_{\min}$, $\theta$ is set to $\theta + 1$. If $\theta < \theta_{\max}$, the algorithm terminates. |

Note: AlphaTruss (Luo et al. 2022a, Luo et al. 2022b); IMCTS = improved Monte Carlo tree search.

## 4. Numerical examples with discrete design variables

In this section, three truss optimization problems with discrete variables are performed to validate the proposed algorithm. The test problems include a 10-bar planar truss with 10 design variables, a 72-bar spatial truss with 16 design variables, and a 220-bar transmission tower with 49 design variables. The first two examples refer to the benchmark problems which have been previously optimized using metaheuristic algorithms. The other one example addresses challenging practical engineering problem. The IMCTS formulation and the direct stiffness method for the analysis of truss structures are implemented in Python software. The computations are carried out with Intel Core i7 2.30 GHz processor and 40 GB memory.

### 4.1. Problem description

Fig. 15 demonstrates the 10-bar planar truss, previously being analyzed by many metaheuristic algorithms, such as GA (Rajeev and Krishnamoorthy 1992), ACO (Camp and Bichon 2004), particle swarm optimizer (PSO) (Li, Huang, and Liu 2009), artificial bee colony (ABC) algorithm (Sonmez 2011), mine blast algorithm (MBA) (Sadollah et al. 2012), water cycle algorithm (WCA) (Eskandar, Sadollah, and Bahreininejad 2013), subset simulation algorithm (SSA) (Li and Ma 2015), and artificial coronary circulation system (ACCS) (Kooshkbaghi, Kaveh, and Zarfam 2020). Various design parameters considered here are shown in Tables 2-4.

The 72-bar spatial truss is shown in Fig. 16, which has been investigated by many methods, such as GA (Wu and Chow 1995), HS (Lee et al. 2005), PSO (Li, Huang, and Liu 2009), MBA (Sadollah et al. 2012), SSA (Li and Ma 2015), and ACCS (Kooshkbaghi, Kaveh, and Zarfam 2020). For analysis, various design parameters are captured from Tables 2-4.

The 220-bar transmission tower is shown in Fig. 17. Similarly, like the previous cases for analysis purpose, various design parameters are taken from Tables 2-4.

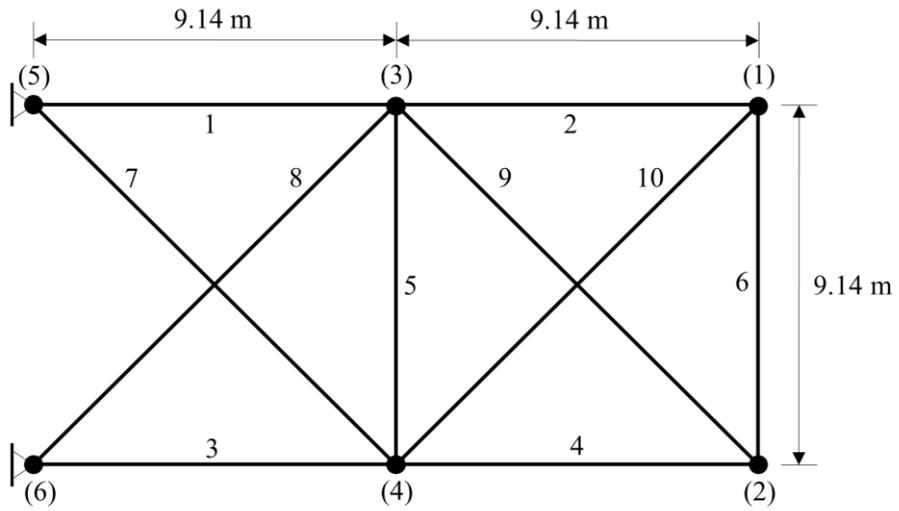

**Fig. 15** A 10-bar planar truss.

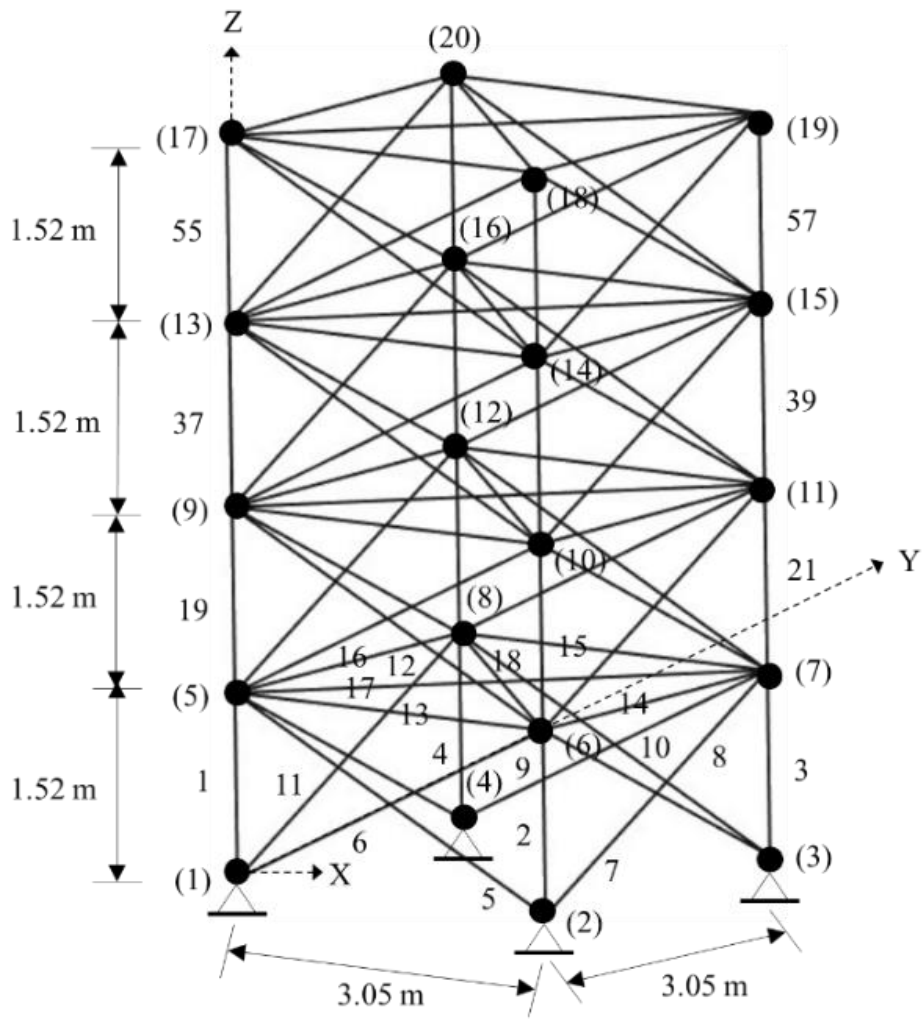

**Fig. 16** A 72-bar spatial truss.

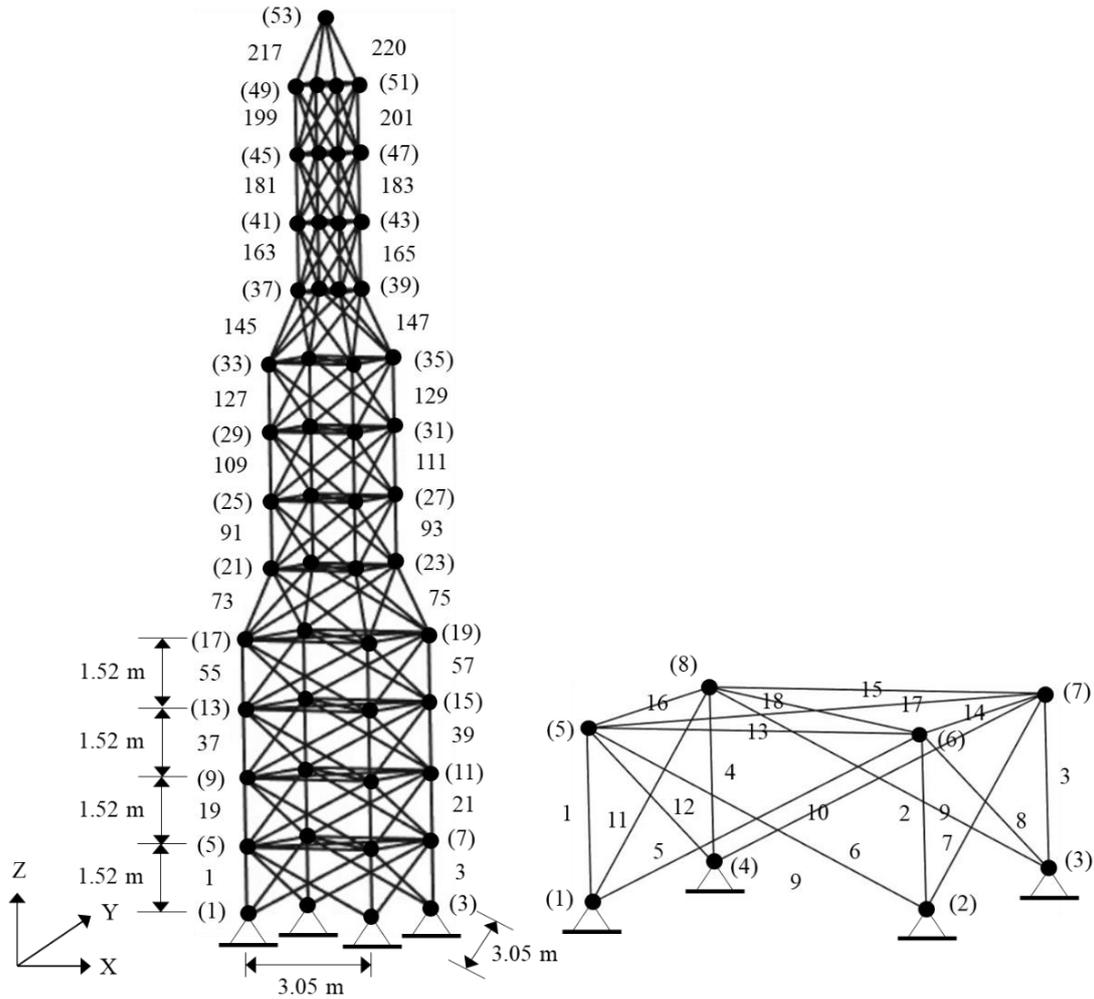

**Fig. 17** A 220-bar transmission tower

**Table 2** Design consideration of the truss optimization problems

| | 10-bar planar truss | 72-bar spatial truss | 220-bar transmission tower |
|---|---|---|---|
| Design variables | $A_i, i = 1, 2, ..., 10$ | $A_i, i = 1, 2, ..., 16$ | $A_i, i = 1, 2, ..., 49$ |
| Material density (kg/m³) | 2767.99 | 2767.99 | 7860.00 |
| Modulus of elasticity (GPa) | 68.95 | 68.95 | 207.00 |
| Stress limitation (MPa) | ±172.37 | ±172.37 | ±180.00 |
| Displacement limitation (mm) | ±50.80 | ±6.35 | ±6.35 |
| Load (kN) | $P_2^y = P_4^y = 444.82$ | Table 5 | Table 6 |

**Table 3** Member group of the truss optimization problems.

| Truss problems | Member group |
|---|---|
| 10-bar planar truss | No member group |
| 72-bar spatial truss | 72 truss members grouped into 16 design variables, as follows: (1) 1–4, (2) 5–12, (3) 13–16, (4) 17–18, (5) 19–22, (6) 23–30, (7) 31–34, (8) 35–36, (9) 37–40, (10) 41–48, (11) 49–52, (12) 53–54, (13) 55–58, (14) 59–66, (15) 67–70, (16) 71–72 |
| 220-bar transmission tower | Table 7 |

**Table 4** Discrete set of the truss optimization problems

| Truss problems | Discrete set |
|---|---|
| 10-bar planar truss | Case 1: $\mathbf{D}$ = {1045.16, 1161.29, 1283.87, 1374.19, 1535.48, 1690.32, 1696.77, 1858.06, 1890.32, 1993.54, 2019.35, 2180.64, 2238.71, 2290.32, 2341.93, 2477.41, 2496.77, 2503.22, 2696.77, 2722.58, 2896.77, 2961.28, 3096.77, 3206.45, 3303.22, 3703.22, 4658.06, 5141.93, 7419.34, 8709.66, 8967.72, 9161.27, 9999.98, 10322.56, 10903.20, 12129.01, 12838.68, 14193.52, 14774.16, 17096.74, 19354.80, 21612.86} mm² <br> Case 2: $\mathbf{D}$ = {64.52, 322.60, 645.20…19678.60, 20001.20, 20323.80} mm² |
| 72-bar spatial truss | Case 1: $\mathbf{D}$ = {64.50, 129.00, 193.50…1935.00, 1999.50, 2064.00} mm² <br> Case 2: American Institute of Steel Construction (AISC) (Table 8) <br> Case 3: $\mathbf{D}$ = {64.50, 129.00, 193.50…1483.50, 1548.00, 1612.50} mm² |
| 220-bar transmission tower | AISC (Table 8) |

**Table 5** Load cases for the 72-bar spatial truss

| Nodes $k$ | Load case 1 (kN) | | | Load case 2 (kN) | | |
|---|---|---|---|---|---|---|
| | $P_k^x$ | $P_k^y$ | $P_k^z$ | $P_k^x$ | $P_k^y$ | $P_k^z$ |
| 17 | 22.24 | 22.24 | −22.24 | 0.00 | 0.00 | −22.24 |
| 18 | 0.00 | 0.00 | 0.00 | 0.00 | 0.00 | −22.24 |
| 19 | 0.00 | 0.00 | 0.00 | 0.00 | 0.00 | −22.24 |
| 20 | 0.00 | 0.00 | 0.00 | 0.00 | 0.00 | −22.24 |

**Table 6** Load cases for the 220-bar transmission tower

| Nodes $k$ | Load case 1 (kN) | | | Load case 2 (kN) | | |
|---|---|---|---|---|---|---|
| | $P_k^x$ | $P_k^y$ | $P_k^z$ | $P_k^x$ | $P_k^y$ | $P_k^z$ |
| 49 | 0.00 | 0.00 | 0.00 | 0.00 | 0.00 | −400.00 |
| 50 | 0.00 | 0.00 | 0.00 | 0.00 | 0.00 | −400.00 |
| 51 | 0.00 | 0.00 | 0.00 | 0.00 | 0.00 | −400.00 |
| 52 | 0.00 | 0.00 | 0.00 | 0.00 | 0.00 | −400.00 |
| 53 | 0.00 | 0.00 | −2000.00 | 0.00 | 0.00 | −400.00 |

**Table 7** Member group for the 220-bar transmission tower

| Number | Members | Number | Members | Number | Members | Number | Members |
|---|---|---|---|---|---|---|---|
| 1 | 1–4 | 14 | 59–66 | 27 | 121–124 | 40 | 179–180 |
| 2 | 5–12 | 15 | 67–70 | 28 | 125–126 | 41 | 181–184 |
| 3 | 13–16 | 16 | 71–72 | 29 | 127–130 | 42 | 185–192 |
| 4 | 17–18 | 17 | 73–76 | 30 | 131–138 | 43 | 193–196 |
| 5 | 19–22 | 18 | 77–84 | 31 | 139–142 | 44 | 197–198 |
| 6 | 23–30 | 19 | 85–88 | 32 | 143–144 | 45 | 199–202 |
| 7 | 31–34 | 20 | 89–90 | 33 | 145–148 | 46 | 203–210 |
| 8 | 35–36 | 21 | 91–94 | 34 | 149–156 | 47 | 211–214 |
| 9 | 37–40 | 22 | 95–102 | 35 | 157–160 | 48 | 215–216 |
| 10 | 41–48 | 23 | 103–106 | 36 | 161–162 | 49 | 217–220 |
| 11 | 49–52 | 24 | 107–108 | 37 | 163–166 | | |
| 12 | 53–54 | 25 | 109–112 | 38 | 167–174 | | |
| 13 | 55–58 | 26 | 113–120 | 39 | 175–178 | | |

**Table 8** Discrete values available for cross-sectional areas from AISC norm

| Number | Area (mm²) | Number | Area (mm²) | Number | Area (mm²) | Number | Area (mm²) |
|---|---|---|---|---|---|---|---|
| 1 | 71.61 | 17 | 1008.39 | 33 | 2477.41 | 49 | 7419.34 |
| 2 | 90.97 | 18 | 1045.16 | 34 | 2496.77 | 50 | 8709.66 |
| 3 | 126.45 | 19 | 1161.29 | 35 | 2503.22 | 51 | 8967.72 |
| 4 | 161.29 | 20 | 1283.87 | 36 | 2696.77 | 52 | 9161.27 |
| 5 | 198.06 | 21 | 1374.19 | 37 | 2722.58 | 53 | 9999.98 |
| 6 | 252.26 | 22 | 1535.48 | 38 | 2896.77 | 54 | 10322.56 |
| 7 | 285.16 | 23 | 1690.32 | 39 | 2961.28 | 55 | 10903.20 |
| 8 | 363.23 | 24 | 1696.77 | 40 | 3096.77 | 56 | 12129.01 |
| 9 | 388.39 | 25 | 1858.06 | 41 | 3206.45 | 57 | 12838.68 |
| 10 | 494.19 | 26 | 1890.32 | 42 | 3303.22 | 58 | 14193.52 |
| 11 | 506.45 | 27 | 1993.54 | 43 | 3703.22 | 59 | 14774.16 |
| 12 | 641.29 | 28 | 2019.35 | 44 | 4658.06 | 60 | 15806.42 |
| 13 | 645.16 | 29 | 2180.64 | 45 | 5141.93 | 61 | 17096.74 |
| 14 | 792.26 | 30 | 2238.71 | 46 | 5503.22 | 62 | 18064.48 |
| 15 | 816.77 | 31 | 2290.32 | 47 | 5999.99 | 63 | 19354.80 |
| 16 | 940.00 | 32 | 2341.93 | 48 | 6999.99 | 64 | 21612.86 |

4.2 Influence of the update process on the optimal solution

The influence of the update process on the optimal solution is investigated for all benchmark problems. Table 9 lists the comparison of optimal solutions obtained from AlphaTruss and IMCTS formulation. It can be observed that IMCTS formulation can find the optimal solution because update process is considered. However, the solution found by AlphaTruss can trap into the local optimum.

**Table 9** Influence of the update process on the optimal solution.

|  | 10-bar planar truss (Case 1) | 10-bar planar truss (Case 2) | 72-bar spatial truss (Case 1) | 72-bar spatial truss (Case 2) |
|---|---|---|---|---|
| AlphaTruss | 2734.23 kg | 3606.47 kg | 192.35 kg | 184.65 kg |
| IMCTS formulation | 2490.56 kg | 2298.50 kg | 174.88 kg | 176.60 kg |

Note: AlphaTruss (Luo et al. 2022a, Luo et al. 2022b); IMCTS = improved Monte Carlo tree search

4.3 Investigation of convergence history: 10-bar planar truss

Fig. 18 illustrates the comparison of convergence histories for the 10-bar planar truss under the best and average reward. Fig. 18(a) and (b) for Case 1 and Case 2 demonstrates that the proposed method obtains the best solution at 13 and 15 rounds under the best reward. However, this method does not detect the best solution after 40 rounds under the average reward.

The comparison of convergence histories for the 10-bar planar truss under different parameter $\alpha$ is shown in Fig. 19. From Fig. 19(a) and 19(b), this algorithm obtains the best solution at 13 and 18 rounds under minimum weight for Case 1 and Case 2.

However, for Case 1 and Case 2, this algorithm does not detect the best solution after 40 rounds under maximum weight.

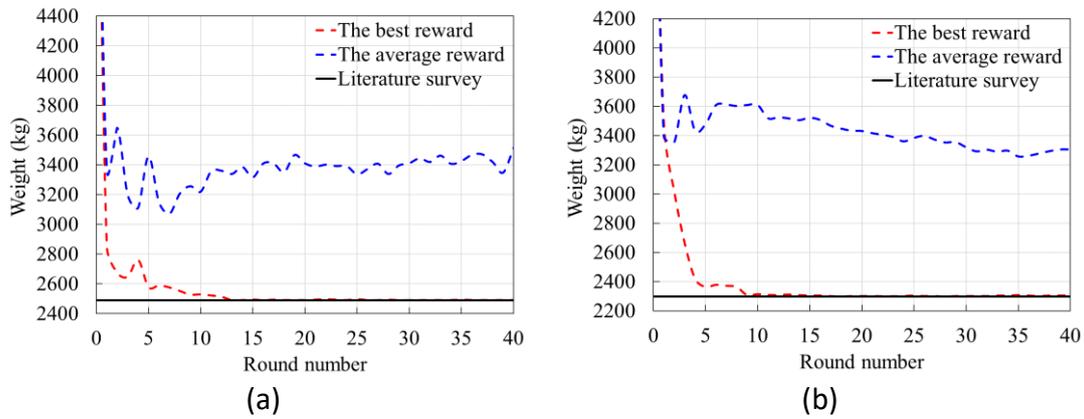

**Fig. 18** Comparison of the convergence histories for 10-bar planar truss under the best and average reward for (a) Case 1 and (b) Case 2.

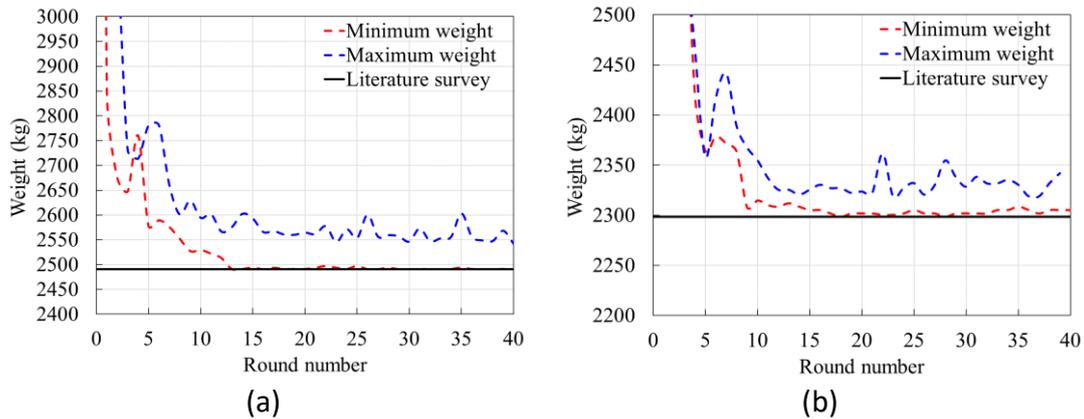

**Fig. 19** Comparison of the convergence histories for 10-bar planar truss under parameter α equal to minimum and maximum weight for (a) Case 1 and (b) Case 2.

### 4.4. Influence of different types of accelerating techniques on the computation time

The influence of accelerating techniques based on the computational effort is investigated for all benchmark problems. Table 10 compares the efficiency of IMCTS without accelerating technique and with different types of accelerating techniques based on CPU time. It can be seen that geometric decay is 1.5, 2.2, and 4.8 times faster than linear decrease, step reduction, and without accelerating technique.

**Table 10** CPU time of IMCTS formulation without and with different types of accelerating techniques

|  | Without accelerating technique | Geometric decay | Linear decrease | Step reduction |
|---|---|---|---|---|
| 10-bar planar truss (Case 1) | 223.49 sec | 46.53 sec | 62.94 sec | 102.47 sec |
| 10-bar planar truss (Case 2) | 380.13 sec | 86.02 sec | 102.45 sec | 264.03 sec |
| 72-bar spatial truss (Case 1) | 12749.90 sec | 2549.98 sec | 3470.55 sec | 4978.25 sec |
| 72-bar spatial truss (Case 2) | 36154.62 sec | 8408.05 sec | 14851.48 sec | 20872.75 sec |

### 4.5. Solution accuracy

Tables 11-14 compare the optimal designs of this method with other algorithms from the literature. It is clear that this method has the ability to provide the best solution (lightest weight) compared with other algorithms. Table 15 is the optimal design for 220-bar transmission tower. Figs. 20-24 reveal that the optimal solution assessed by this approach does not violate stress and displacement constraints. The results demonstrate that this method is suitable for benchmark and large-scale truss structures.

**Table 11.** Comparison of optimum designs for the 10-bar planar truss (Case 1).

| Group number | Members | Optimal cross-sectional areas (mm$^2$) | | | | | |
|---|---|---|---|---|---|---|---|
|  |  | GA | ACO | PSO | ABC | ACCS | IMCTS |
| 1 | 1 | 21612.86 | 21612.86 | 19354.80 | 21612.86 | 21612.86 | 21612.86 |
| 2 | 2 | 1045.16 | 1045.16 | 1045.16 | 1045.16 | 1045.16 | 1045.16 |
| 3 | 3 | 14193.52 | 14774.16 | 19354.80 | 14774.16 | 14774.16 | 14774.16 |
| 4 | 4 | 9999.98 | 9161.27 | 8709.66 | 9161.27 | 9161.27 | 9161.27 |
| 5 | 5 | 1045.16 | 1045.16 | 1045.16 | 1045.16 | 1045.16 | 1045.16 |
| 6 | 6 | 1045.16 | 1045.16 | 1161.29 | 1045.16 | 1045.16 | 1045.16 |
| 7 | 7 | 9161.27 | 5141.93 | 7419.34 | 5141.93 | 5141.93 | 5141.93 |
| 8 | 8 | 12838.68 | 14774.16 | 12129.01 | 14774.16 | 14774.16 | 14774.16 |
| 9 | 9 | 12838.68 | 14193.52 | 14193.52 | 14193.52 | 14193.52 | 14193.52 |
| 10 | 10 | 1690.32 | 1045.16 | 1161.29 | 1045.16 | 1045.16 | 1045.16 |
| Optimal weight (kg) |  | 2546.39 | 2490.56 | 2531.84 | 2490.56 | 2490.56 | 2490.56 |

Note: GA = genetic algorithm (Rajeev and Krishnamoorthy 1992); ACO = ant colony optimization (Camp and Bichon 2004); PSO = particle swarm optimizer (Li, Huang, and Liu 2009); ABC = artificial bee colony (Sonmez 2011); ACCS = artificial coronary circulation system (Kooshkbaghi, Kaveh, and Zarfam 2020); IMCTS = improved Monte Carlo tree search.

**Table 12.** Comparison of optimum designs for the 10-bar planar truss (Case 2).

| Group number | Members | Optimal cross-sectional areas (mm$^2$) | | | | | |
|---|---|---|---|---|---|---|---|
| | | PSO | MBA | WCA | SSA | ACCS | IMCTS |
| 1 | 1 | 15807.40 | 19033.40 | 19356.00 | 19356.00 | 19678.60 | 19678.60 |
| 2 | 2 | 64.52 | 64.52 | 64.52 | 64.52 | 64.52 | 64.52 |
| 3 | 3 | 14517.00 | 15484.80 | 15162.20 | 15162.20 | 14839.60 | 15162.20 |
| 4 | 4 | 10000.60 | 9678.00 | 9678.00 | 9678.00 | 9678.00 | 9355.40 |
| 5 | 5 | 64.52 | 64.52 | 64.52 | 64.52 | 64.52 | 64.52 |
| 6 | 6 | 967.80 | 322.60 | 322.60 | 322.60 | 322.60 | 322.60 |
| 7 | 7 | 5484.20 | 4839.00 | 4839.00 | 4839.00 | 4839.00 | 4839.00 |
| 8 | 8 | 13871.80 | 13871.80 | 13871.80 | 13871.80 | 13549.20 | 13549.20 |
| 9 | 9 | 17743.00 | 13871.80 | 13871.80 | 13871.80 | 14194.40 | 14194.40 |
| 10 | 10 | 64.52 | 64.52 | 64.52 | 64.52 | 64.52 | 64.52 |
| Optimal weight (kg) | | 2378.51 | 2298.50 | 2298.50 | 2298.50 | 2298.50 | 2298.50 |

Note: PSO = particle swarm optimizer (Li, Huang, and Liu 2009); MBA = mine blast algorithm (Sadollah et al. 2012); WCA = water cycle algorithm (Eskandar, Sadollah, and Bahreininejad 2013); SSA = subset simulation algorithm (Li and Ma 2015); ACCS = artificial coronary circulation system (Kooshkbaghi, Kaveh, and Zarfam 2020); IMCTS = improved Monte Carlo tree search.

Table 13. Comparison of optimum designs for the 72-bar spatial truss (Case 1).

| Group number | Members | Optimal cross-sectional areas (mm$^2$) | | | | | |
|---|---|---|---|---|---|---|---|
| | | GA | HS | PSO | MBA | ACCS | IMCTS |
| 1 | 1–4 | 967.50 | 1225.50 | 1677.00 | 1290.00 | 1290.00 | 1290.00 |
| 2 | 5–12 | 451.50 | 322.50 | 967.50 | 387.00 | 322.50 | 322.50 |
| 3 | 13–16 | 64.50 | 64.50 | 193.50 | 258.00 | 64.50 | 64.50 |
| 4 | 17–18 | 64.50 | 64.50 | 64.50 | 387.00 | 64.50 | 64.50 |
| 5 | 19–22 | 838.50 | 903.00 | 1354.50 | 322.50 | 838.50 | 838.50 |
| 6 | 23–30 | 322.50 | 387.00 | 967.50 | 322.50 | 322.50 | 322.50 |
| 7 | 31–34 | 129.00 | 64.50 | 387.00 | 64.50 | 64.50 | 64.50 |
| 8 | 35–36 | 64.50 | 64.50 | 193.50 | 64.50 | 64.50 | 64.50 |
| 9 | 37–40 | 322.50 | 387.00 | 1419.00 | 903.00 | 322.50 | 322.50 |
| 10 | 41–48 | 322.50 | 322.50 | 1225.50 | 322.50 | 322.50 | 322.50 |
| 11 | 49–52 | 64.50 | 64.50 | 129.00 | 64.50 | 64.50 | 64.50 |
| 12 | 53–54 | 129.00 | 64.50 | 580.50 | 64.50 | 64.50 | 64.50 |
| 13 | 55–58 | 129.00 | 129.00 | 258.00 | 1225.50 | 129.00 | 129.00 |
| 14 | 59–66 | 322.50 | 322.50 | 1225.50 | 322.50 | 387.00 | 387.00 |
| 15 | 67–70 | 322.50 | 258.00 | 451.50 | 64.50 | 258.00 | 258.00 |
| 16 | 71–72 | 451.50 | 387.00 | 1032.00 | 64.50 | 387.00 | 387.00 |
| Optimal weight (kg) | | 181.74 | 175.97 | 494.36 | 174.88 | 174.88 | 174.88 |

Note: GA = genetic algorithm (Wu and Chow 1995); HS = harmony search (Lee et al. 2005); PSO = particle swarm optimizer (Li, Huang, and Liu 2009); MBA = mine blast algorithm (Sadollah et al. 2012); ACCS = artificial coronary circulation system (Kooshkbaghi, Kaveh, and Zarfam 2020); IMCTS = improved Monte Carlo tree search.

Table 14. Comparison of optimum designs for the 72-bar spatial truss (Case 2).

| Group number | Members | Optimal cross-sectional areas (mm²) | | | | | |
|---|---|---|---|---|---|---|---|
| | | GA | PSO | MBA | SSA | ACCS | IMCTS |
| 1 | 1–4 | 126.45 | 4658.06 | 126.45 | 1283.87 | 1283.87 | 1283.87 |
| 2 | 5–12 | 388.39 | 1161.29 | 363.23 | 363.23 | 285.16 | 363.23 |
| 3 | 13–16 | 198.06 | 729.03 | 285.16 | 71.61 | 71.61 | 71.61 |
| 4 | 17–18 | 494.19 | 126.45 | 388.39 | 71.61 | 71.61 | 71.61 |
| 5 | 19–22 | 252.26 | 1993.54 | 285.16 | 792.26 | 792.26 | 792.26 |
| 6 | 23–30 | 252.26 | 506.45 | 285.16 | 363.23 | 363.23 | 285.16 |
| 7 | 31–34 | 90.97 | 363.23 | 71.61 | 71.61 | 71.61 | 71.61 |
| 8 | 35–36 | 71.61 | 506.45 | 71.61 | 71.61 | 71.61 | 71.61 |
| 9 | 37–40 | 1161.29 | 1993.54 | 816.77 | 363.23 | 363.23 | 363.23 |
| 10 | 41–48 | 388.39 | 792.26 | 363.23 | 285.16 | 363.23 | 363.23 |
| 11 | 49–52 | 90.97 | 71.61 | 71.61 | 71.61 | 71.61 | 71.61 |
| 12 | 53–54 | 198.06 | 363.23 | 71.61 | 71.61 | 71.61 | 71.61 |
| 13 | 55–58 | 1008.39 | 1283.87 | 1161.29 | 126.45 | 126.45 | 126.45 |
| 14 | 59–66 | 494.19 | 1045.16 | 388.39 | 363.23 | 363.23 | 363.23 |
| 15 | 67–70 | 90.97 | 1008.39 | 71.61 | 252.26 | 252.26 | 252.26 |
| 16 | 71–72 | 71.61 | 816.77 | 71.61 | 363.23 | 363.23 | 363.23 |
| Optimal weight (kg) | | 193.78 | 548.61 | 177.23 | 176.60 | 176.60 | 176.60 |

Note: GA = genetic algorithm (Wu and Chow 1995); PSO = particle swarm optimizer (Li, Huang, and Liu 2009); MBA = mine blast algorithm (Sadollah et al. 2012); SSA = subset simulation algorithm (Li and Ma 2015); ACCS = artificial coronary circulation system (Kooshkbaghi, Kaveh, and Zarfam 2020); IMCTS = improved Monte Carlo tree search.

**Table 15** Optimal design for 220-bar transmission tower

| Number | Area (mm²) | Number | Area (mm²) | Number | Area (mm²) | Number | Area (mm²) |
|---|---|---|---|---|---|---|---|
| 1 | 8967.72 | 14 | 71.61 | 27 | 90.97 | 40 | 363.23 |
| 2 | 71.61 | 15 | 2238.71 | 28 | 506.45 | 41 | 8967.24 |
| 3 | 71.61 | 16 | 1008.39 | 29 | 9161.27 | 42 | 71.61 |
| 4 | 71.61 | 17 | 9999.98 | 30 | 71.61 | 43 | 71.61 |
| 5 | 8967.72 | 18 | 71.61 | 31 | 2290.32 | 44 | 71.61 |
| 6 | 71.61 | 19 | 1535.48 | 32 | 1008.39 | 45 | 9161.27 |
| 7 | 71.61 | 20 | 2496.77 | 33 | 10322.56 | 46 | 71.61 |
| 8 | 71.61 | 21 | 9161.27 | 34 | 71.61 | 47 | 1535.48 |
| 9 | 8709.66 | 22 | 71.61 | 35 | 645.16 | 48 | 1690.32 |
| 10 | 71.61 | 23 | 90.97 | 36 | 2961.28 | 49 | 10322.56 |
| 11 | 71.61 | 24 | 161.29 | 37 | 8967.24 | | |
| 12 | 71.61 | 25 | 9161.27 | 38 | 90.97 | | |
| 13 | 8967.72 | 26 | 90.97 | 39 | 71.61 | | |

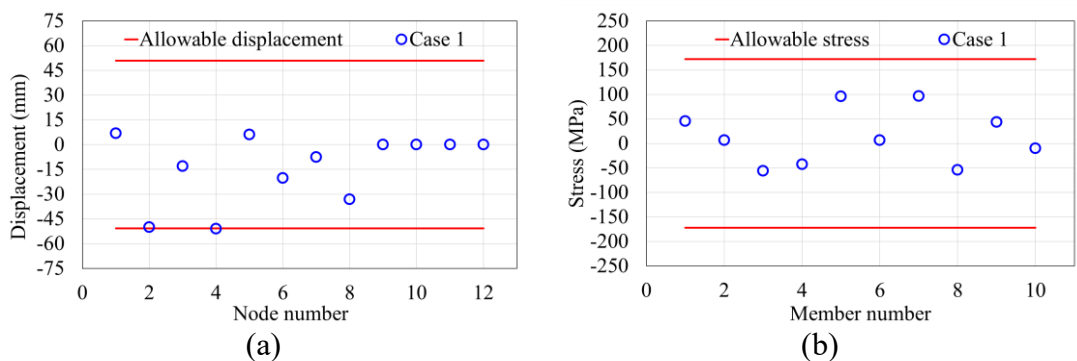

(a)      (b)

**Fig. 20** Constraints evaluated at the optimal design of 10-bar planar truss (Case 1) by

IMCTS formulation for (a) displacement constraints and (b) stress constraints.

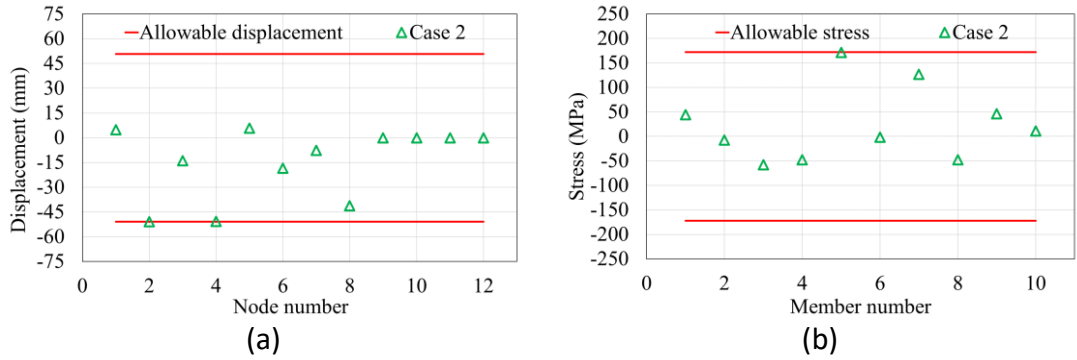

(a) (b)

**Fig. 21** Constraints evaluated at the optimal design of 10-bar planar truss (Case 2) by IMCTS formulation for (a) displacement constraints and (b) stress constraints

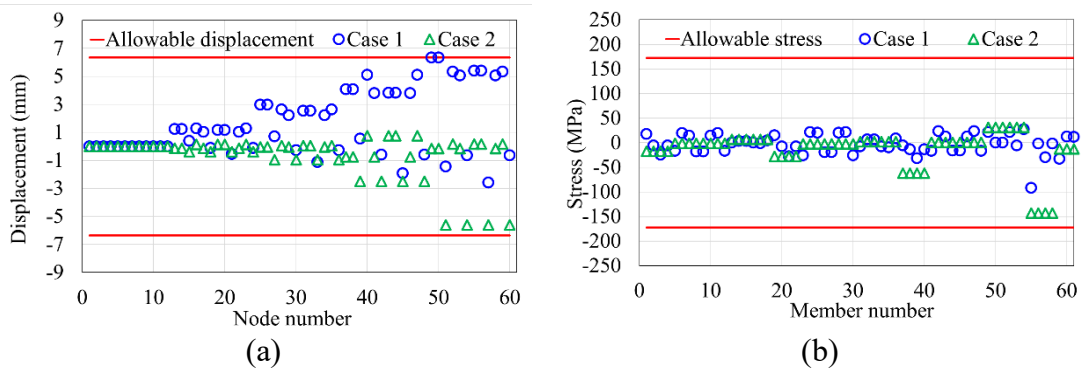

(a) (b)

**Fig. 22** Constraints evaluated at the optimal design of 72-bar spatial truss (Case 1) by IMCTS formulation for (a) displacement constraints and (b) stress constraints

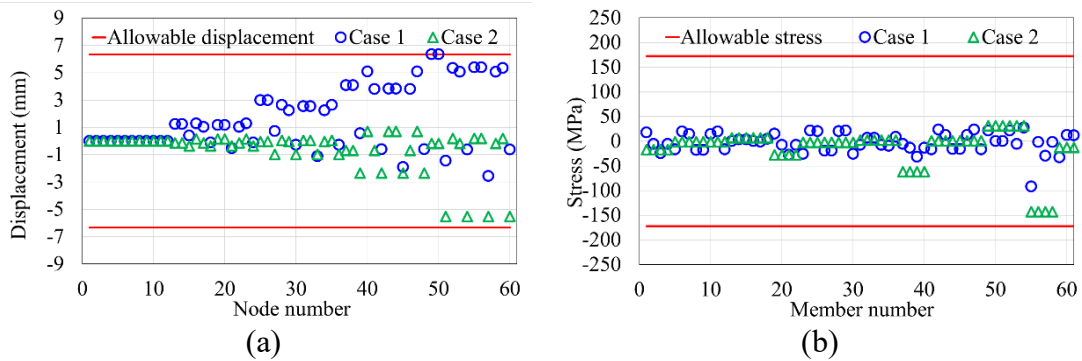

(a) (b)

**Fig. 23** Constraints evaluated at the optimal design of 72-bar spatial truss (Case 2) by IMCTS formulation for (a) displacement constraints and (b) stress constraints

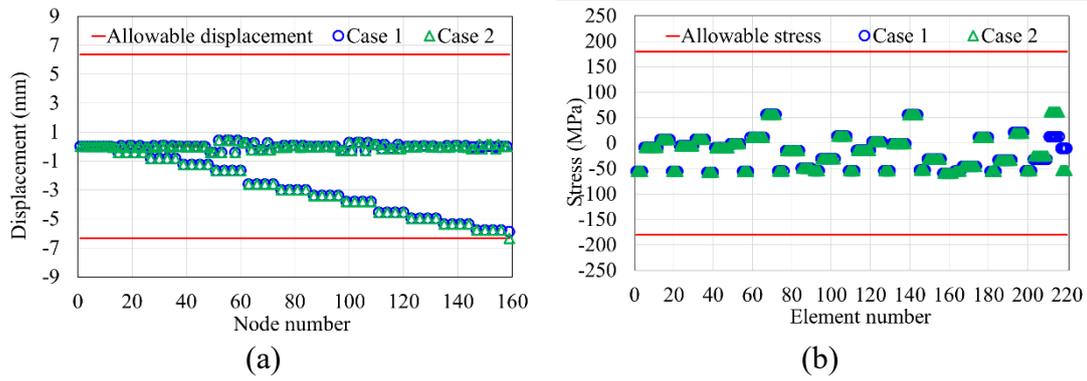

(a)            (b)

**Fig. 24** Constraints evaluated at the optimal design of 220-bar transmission tower by IMCTS formulation for (a) displacement constraints and (b) stress constraints

4.6 Applicability of multi-objective structure optimization: comparison with metaheuristic algorithms

The Pareto front is widely used in multi-objective optimization. For truss optimization problems, the Pareto front presents the relationship between weight and maximum displacement. Fig. 25(a) and (b) shows Pareto fronts from the literature (Kumar et al. 2023) and IMCTS formulation for 10-bar planar truss (Case 1) and 72-bar spatial truss (Case 3). It is seen that the Pareto front of IMCTS formulation is similar to that of multi-objective multi-verse optimizer (MOMVO).

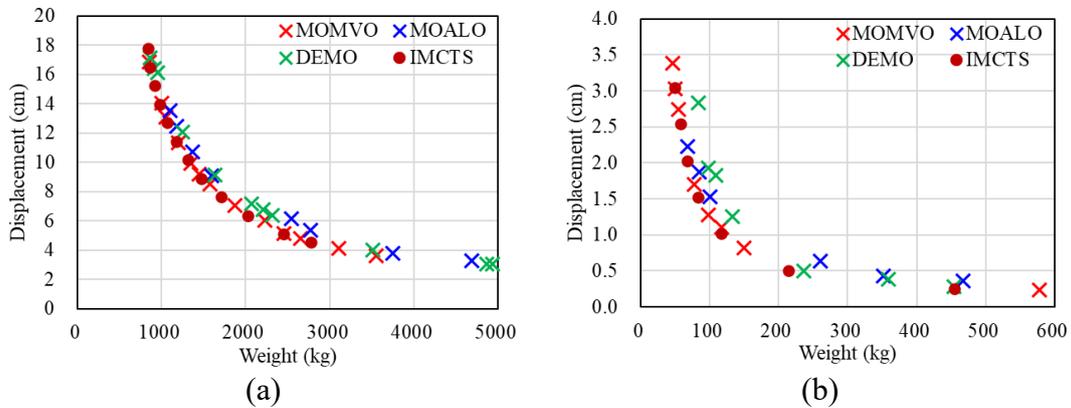

(a)            (b)

**Fig. 25** Pareto fronts from metaheuristic algorithms and IMCTS formulation for (a) 10-bar planar truss (Case 1) and (b) 72-bar spatial truss (Case 3). MOMVO = multi-objective multi-verse optimizer (Kumar et al. 2023); MOALO = MO ant lion optimizer (Mirjalili, Jangir, and Saremi 2017); DEMO = MO differential evolution (Mirjalili 2016); IMCTS = improved Monte Carlo tree search

4.7 Solution stability

The statistical results of all truss optimization problems are obtained through 10 independent sampling to test the stability of this method. The results are summarized in Table 16, including the best, the worst, average, and standard deviation.

**Table 16** Statistical results of the investigated example

| Investigated example | Best weight | Worst weight | Average weight | Standard deviation |
|---|---|---|---|---|
| 10-bar planar truss (Case 1) | 2490.56 kg | 2523.38 kg | 2500.94 kg | 11.81 |
| 10-bar planar truss (Case 2) | 2298.50 kg | 2308.07 kg | 2303.26 kg | 3.15 |
| 72-bar spatial truss (Case 1) | 174.88 kg | 177.06 kg | 176.07 kg | 0.60 |
| 72-bar spatial truss (Case 2) | 176.60 kg | 179.47 kg | 176.94 kg | 0.89 |
| 220-bar transmission tower | 6991.00 kg | 7193.25 kg | 7091.14 kg | 64.73 |

## 5. Concluding remarks

In this paper, a novel RL algorithm using IMCTS formulation with multiple root nodes has been formulated for sizing optimization of truss structures. The cross-sectional areas of members are selected from a prescribed list of discrete values to minimize the weight of the truss under stress and displacement constraints. It is difficult for this problem to find an optimal solution in a reasonable time. This algorithm incorporates update process, the best reward, accelerating technique, and terminal condition. The following are the key points of this research:

(1) Once a design variable vector is determined by search tree in this round, this solution is used as the solution for initial state in next round. The above-mentioned process is called an update process.

(2) The best reward is used in the backpropagation step, which is defined as the maximum value between the current state value and the result of simulation.

(3) Accelerating technique is developed by decreasing the width of search tree as the update process proceeds and reducing maximum number of iterations during policy improvement process.

(4) Three types of accelerating techniques including geometric decay, linear decrease, and step reduction are considered in this study. Geometric decay performs better than linear decrease and step reduction in relation with computational cost.

(5) In each round, the agent is trained to find the solution under various constraints until the terminal condition is satisfied. When the improvement factor is less than 0.01%, the counter increases by 1. When the counter is greater or equal to 3, the algorithm terminates. Then, optimal solution is the minimum value of all solutions found by search trees during the algorithm.

The proposed algorithm allows the agent to find the optimal solution. The numerical examples demonstrate that this algorithm provides results as good as metaheuristic algorithms with very low computational cost and applicable for practical engineering problems and multi-objective structural optimization. In conclusion, this study suggests that the novel RL-based algorithm using IMCTS formulation is a powerful optimization technique for discrete truss optimization without applying parameter-based search mechanisms. When compared to Q-learning-based RL method and AlphaTruss, IMCTS formulation can be applied to optimization problems with discrete variables and requires less computational effort for the training.

The proposed algorithm is expected to solve optimization problems with multiple types of variables considering single and mixed continuous-discrete variables, which are our future research interests.